%% file: main.tex
\documentclass[letterpaper]{article} 
\usepackage[preprint]{aaai2027}  
\usepackage[hyphens]{url}  
\usepackage{graphicx} 
\usepackage{natbib}  
\usepackage{caption} 
\input{math_commands.tex}

\usepackage[utf8]{inputenc} 
\usepackage[T1]{fontenc}    
\usepackage{booktabs}       
\usepackage{nicefrac}       
\usepackage{microtype}      
\usepackage{multirow}
\usepackage[table]{xcolor}
\usepackage{amssymb}
\usepackage{tabularx}
\usepackage{pifont}
\usepackage[ruled,vlined,linesnumbered]{algorithm2e}
\usepackage{arydshln}
\usepackage{makecell}
\definecolor{darkgreen}{RGB}{34, 139, 34}
\usepackage{xcolor}
\definecolor{GSPOrange}{HTML}{E97832}
\definecolor{RCRPurple}{HTML}{8054C7}

\newcommand{\tablestyle}[2]{\setlength{\tabcolsep}{#1}\renewcommand{\arraystretch}{#2}\centering\footnotesize}

\title{ACE: Adaptive Calibration-Free Expert Skipping for MoE-based LLMs}
\author{
    Zukang Xu\equalcontrib,
    Zhixiong Zhao\equalcontrib,
    Xing Hu,
    Jiangyong Yu,
    Houji Wen,
    Jun Li,
    Zhe Jiang,
    Dawei Yang\corresponding
}
\affiliations{
    dawei.yang@houmo.ai
}

\begin{document}

\maketitle

\begin{abstract}
    \input{Section/0_abstract}
\end{abstract}

\input{Section/1_intro}

\input{Section/2_related-work}

\input{Section/4_method}
\input{Section/5_experiment}
\input{Section/6_conclusion}

\clearpage
\bibliography{aaai2027}

\clearpage
\appendix
\input{Section/07_appendix}

\end{document}

%% file: math_commands.tex
\usepackage{amsmath,amsfonts,bm}

\def\1{\bm{1}}

\DeclareMathAlphabet{\mathsfit}{\encodingdefault}{\sfdefault}{m}{sl}
\SetMathAlphabet{\mathsfit}{bold}{\encodingdefault}{\sfdefault}{bx}{n}



%% file: Section/0_abstract.tex
Mixture-of-Experts (MoE) architectures provide an efficient paradigm for scaling large language models (LLMs), yet fixed top-$k$ routing executes the same number of expert slots for every token and can retain substantial redundant computation. 
Existing expert-skipping methods often rely on router confidence, calibration data, or additional training,and therefore cannot reliably estimate the actual contribution of routed experts. 
To this end, we propose ACE, a training-free, calibration-free, and checkpoint-preserving framework for token-adaptive expert skipping  in MoE-based LLMs. 
ACE contains two complementary components: 1) Global Spectral Proxy (GSP), which estimates global transformation capacity from the coupled gate, up, and down projections together with  RMSNorm scaling; and 2)  Router-Conditioned Refinement (RCR),which constructs expert-specific direction prototypes from centered router weights and evaluates expert responses along routing-preferred directions. 
During inference, ACE combines both estimates with runtime router gates and skips an expert slot only when both views identify it as low-contribution, while always retaining the top-1 expert. 
All expert statistics are computed offline, leaving only table lookups and lightweight scalar operations online. 
Extensive experiments across three MoE-based LLMs and eight benchmarks demonstrate that ACE consistently outperforms existing static and dynamic baselines, with increasingly pronounced advantages under aggressive expert skipping. 
For instance, at a 50\% skipping ratio on Qwen3.6-35B-A3B, ACE reduces WikiText-2 perplexity by 7.96\% and improves average downstream accuracy by 4.15 percentage points over the strongest competing method.
Optimized expert dispatch further yields up to 2.25$\times$ prefill and 1.41$\times$ decoding speedups in our measurements.
The code is available at \url{https://github.com/xzktx003/moe-prune}.

%% file: Section/1_intro.tex
\section{Introduction}
\label{sec:intro}

Mixture-of-Experts (MoE) has become a key approach for scaling large language models (LLMs)~\citep{shazeer2017outrageously}. Unlike dense Transformers that execute the full feed-forward network at every layer, MoE models replace each FFN sublayer with multiple experts and route each token to only a small subset, thereby expanding model capacity while controlling per-token computation. Recent models further demonstrate this advantage: Qwen3.6-35B-A3B~\citep{qwen36modelcard} activates only about 3B of its 35B parameters, while DeepSeek-V4-Pro and DeepSeek-V4-Flash~\citep{xu2026deepseek} adopt total-to-active parameter configurations of 1.6T/49B and 284B/13B, respectively. However, sparse activation does not make MoE inference inherently efficient. Most MoE-based LLMs still use fixed top-$k$ routing, activating the same number of expert slots for every token despite substantial variation in their contributions. As a result, low-contribution experts may still be fully executed, causing redundant computation and notable inference overhead, especially in resource-constrained or memory-bandwidth-limited deployments.

\begin{figure}[t]
    \centering
    \includegraphics[width=0.9\linewidth]{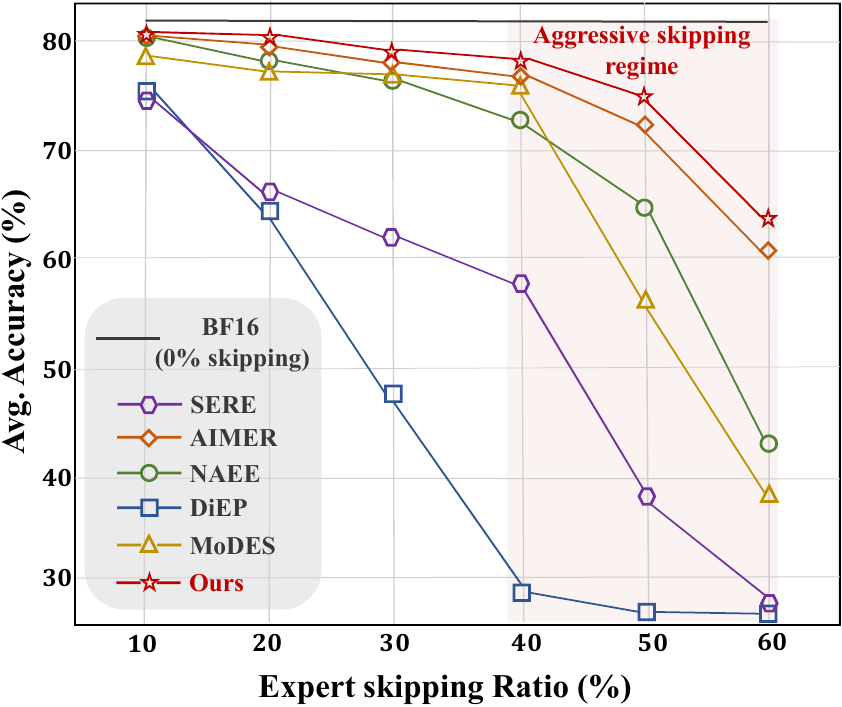}
    \vspace{-0.1cm}
    \caption{Accuracy--efficiency trade-off on Qwen3-30B-A3B-Instruct-2507 under different expert skipping ratios. ACE consistently outperforms all baselines across the full range.}
    \vspace{-0.2cm}
    \label{fig1}
\end{figure}

To further improve MoE inference efficiency, existing studies mainly reduce expert redundancy through offline expert compression or online dynamic activation control. Offline pruning and merging methods~\citep{liu2026aimer,bai2026diep} modify the expert pool before deployment based on expert usage, activation responses, gradient/Fisher information, or search-based criteria, thereby reducing model size and resident memory. 
However, such static compression shares the same expert structure across all inputs and cannot adapt to token-specific demands for routed expert slots. Even after the expert pool is compressed, the remaining MoE layers typically retain fixed top-$k$ routing and execute a preset number of experts, leaving dynamic redundancy within the routed set unaddressed. 
Dynamic expert skipping is more closely aligned with our goal, as it adaptively reduces executed expert slots during inference. Some methods modify or retrain the router~\citep{huang2024harder}, or introduce additional decision modules to learn token-wise expert budgets~\citep{yue2024ada}, while post-training approaches determine skipping thresholds from router-score imbalance, calibration-based layer sensitivity, or activation statistics~\citep{huang2026modes}.
Although these methods avoid permanently removing experts, their decisions still rely largely on router confidence or calibration-dependent signals. The former reflects relative routing preference rather than the effective transformation contribution of an executed expert, whereas the latter requires additional calibration-time forward passes and may be sensitive to calibration size, domain distribution, and task shift. 
As shown in Figure~\ref{fig1}, existing methods remain stable under mild skipping but degrade rapidly as the skipping ratio increases, whereas ACE maintains a superior accuracy--efficiency trade-off in the aggressive 40\%--60\% regime. This suggests that the key challenge of high-ratio skipping is not merely to reduce expert executions, but to reliably estimate the relative contributions of routed expert slots for token-adaptive skipping without calibration data or checkpoint modification.

To understand the key challenges of calibration-free expert-slot skipping, we examine the problem from three perspectives: routing preference, expert structure, and input directionality. 
First, router gates only reflect the relative assignment preference among the top-$k$ candidate experts, rather than the actual contribution of each routed expert slot to the MoE output. Experts with similar gate values may produce substantially different responses due to variations in their FFN structures, RMSNorm scaling, and output mappings; conversely, an expert with a smaller gate may still contribute non-negligibly through stronger transformation capacity. Therefore, skipping experts solely based on router scores may incorrectly equate routing preference with expert contribution. Second, estimating expert transformation capacity directly from model parameters, without calibration activations or additional training, is non-trivial. The output of a SwiGLU expert arises from the nonlinear interaction among its gate, up, and down projections, which cannot be adequately characterized by a single matrix norm, usage frequency, or global routing statistic. Finally, expert contribution is inherently direction-dependent. An expert with weak global responses may still exhibit strong behavior within the local input region favored by the router. A reliable data-free contribution estimator should therefore capture both the global structural capacity of each expert and the directional specialization encoded by the router.

Motivated by these observations, we propose ACE (\textbf{A}daptive \textbf{C}alibration-Free \textbf{E}xpert Skipping), a training-free, calibration-free, and checkpoint-preserving framework for dynamic expert skipping in MoE-based LLMs. 
ACE retains the pretrained router and its original top-$k$ candidate set, while estimating the relative contribution of each routed expert slot by combining token-specific routing preference with expert-specific transformation capacity. Specifically, the Global Spectral Proxy (GSP) extracts a direction-agnostic structural response from the gate, up, and down projections of each SwiGLU expert together with the preceding RMSNorm scaling, thereby characterizing its global transformation capacity. 
Router-Conditioned Refinement (RCR) further constructs an expert-specific direction prototype from the centered router weights and measures the expert response along its preferred routing direction, compensating for directional specialization overlooked by the global proxy. 
During inference, ACE combines the router gates with the precomputed GSP and RCR amplification tables, and progressively skips low-contribution slots under dual-view contribution constraints and a router-mass safeguard, while always preserving the top-1 expert. 
Since all expert-level statistics are derived offline from model weights and router geometry, ACE requires no real samples, activation caching, or threshold profiling, and introduces only table lookups and lightweight top-$k$ scalar operations at runtime. 
Our key contributions can be summarized as follows:
\begin{itemize}
    \item We identify key challenges in calibration-free expert-slot skipping: mismatch between routing preference and expert contribution, structural coupling within SwiGLU experts, and direction-dependent specialization of responses.
    
    \item We propose \textbf{ACE}, a training-free and checkpoint-preserving framework that combines calibration-free GSP and RCR contribution estimates for safeguarded token-adaptive skipping.
    
    \item Extensive experiments across MoE-based LLMs and tasks show that ACE consistently outperforms static and dynamic baselines, especially under aggressive skipping.
\end{itemize}

%% file: Section/2_related-work.tex
\section{Related Work}
\label{sec:related}
\paragraph{Static Expert Compression in MoE-based LLMs.}
Static expert compression reduces model size and resident memory by pruning, merging, or restructuring experts before deployment. Existing methods identify redundant experts using routing frequency, task statistics, or structural importance. The expert-pruning branch of NAEE~\citep{lu2024not} selects fixed subsets for task-agnostic and task-specific compression, EAC-MoE~\citep{chen2025eac} removes rarely selected experts based on selection frequency, and AIMER~\citep{liu2026aimer} develops a calibration-free criterion for task-agnostic pruning. More fine-grained methods include MoE-Pruner~\citep{xie2024moe}, which combines router information with input activations for one-shot intra-expert pruning, and REAP~\citep{lasby2025reap}, which jointly considers router gates and expert activation norms. DiEP~\citep{bai2026diep} learns non-uniform layer-wise sparsity through differentiable optimization, COMPEL~\citep{yoon2026compel} integrates Fisher-based importance, layer-wise outlier distributions, and weight compensation, while STUN~\citep{lee2025stun} applies structured expert pruning followed by unstructured compression within retained experts. Despite reducing parameter and memory costs, these methods determine a fixed structure before inference and apply it uniformly to all inputs. The remaining MoE layers generally retain fixed top-$k$ routing and cannot adaptively eliminate low-contribution expert slots for individual tokens.

\begin{figure*}[ht!]
    \centering
    \includegraphics[width=0.9\textwidth]{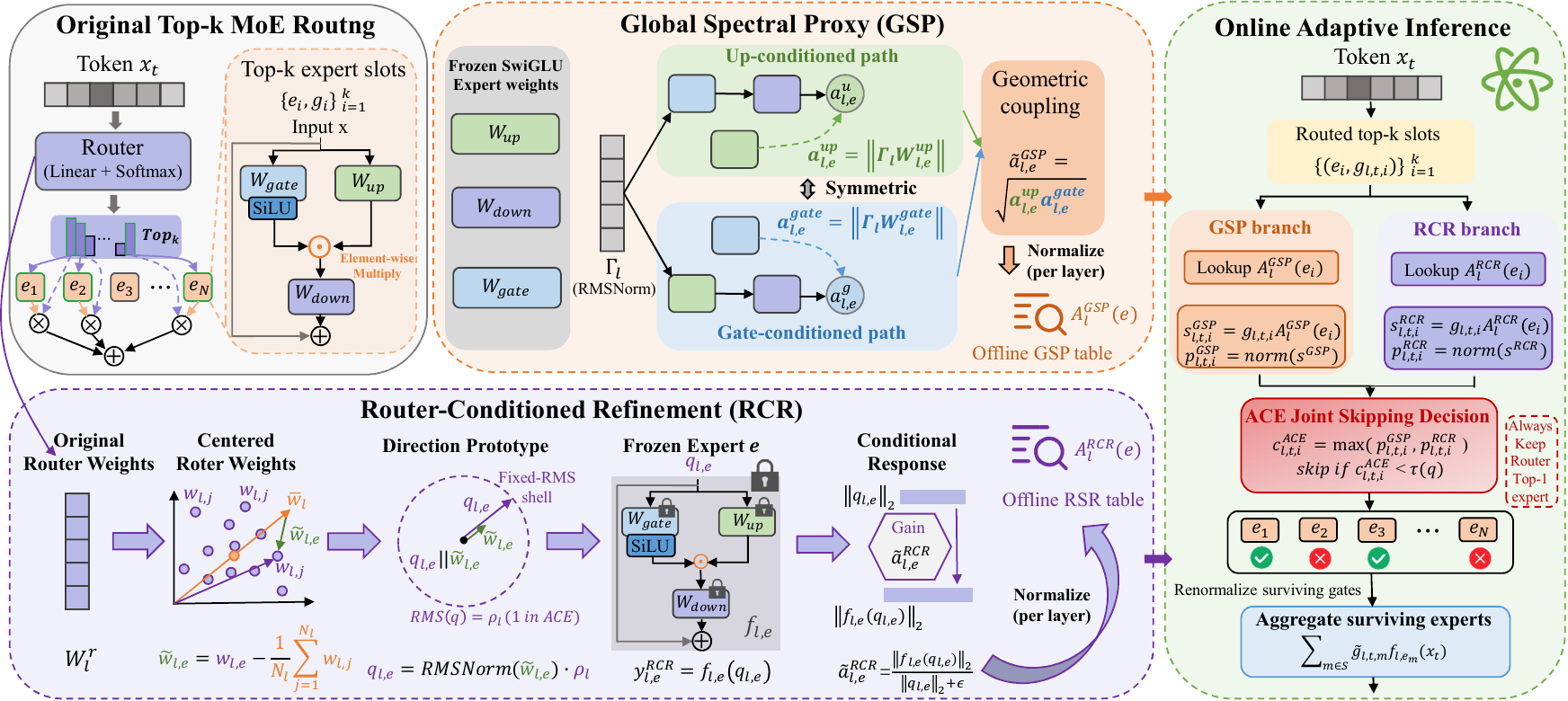}
    \caption{\textbf{Overview of ACE.} 
\textcolor{GSPOrange}{Global Spectral Proxy (GSP)} estimates each expert's transformation capacity from frozen SwiGLU weights, while 
\textcolor{RCRPurple}{Router-Conditioned Refinement (RCR)} evaluates responses along router-induced directions. During inference, ACE combines offline scores with token-wise router gates to skip low-contribution expert slots and aggregate surviving experts.}
    \label{fig:overview}
\end{figure*}

\paragraph{Dynamic Expert Skipping in MoE-based LLMs.}
Dynamic expert skipping reduces expert activations according to token-, layer-, or batch-specific states. Some methods replace fixed top-$k$ routing with variable-cardinality policies. Top-$P$ Routing~\citep{huang2024harder} determines activated expert counts from cumulative routing probabilities, Ada-K Routing~\citep{yue2024ada} learns a lightweight allocator through reinforcement learning, and DynMoE~\citep{guo2025dynamic} combines Top-Any gating with adaptive training to vary expert counts across tokens. For pretrained MoE-based LLMs, NAEE~\citep{lu2024not} skips secondary experts using relative router scores and calibration-derived layer thresholds. Alloc-MoE~\citep{liu2026alloc} distributes a global activation budget through layer sensitivity profiling and dynamic programming, followed by routing-score-based token allocation, while MoDES~\citep{huang2026modes} combines globally modulated local gating with modality-specific threshold search for training-free skipping in multimodal MoE models. Batch-level methods include SERE~\citep{wu2026sere}, which re-routes tokens from secondary to similar primary experts, and XShare~\citep{vankov2026xshare}, which selects shared expert subsets from batch routing demands. Despite their effectiveness, these methods generally require routing-policy training, additional modules, calibration or sensitivity profiling, router-only importance signals, or specific batch-serving assumptions.

%% file: Section/4_method.tex
\section{Method}
\label{sec:method}

\paragraph{Overview.}
To address the three challenges of calibration-free expert skipping identified in Sec.~\ref{sec:intro}, we propose ACE, a parameter-driven framework for token-adaptive expert skipping in MoE-based LLMs (Fig.~\ref{fig:overview}). First, Global Spectral Proxy (GSP) models the multiplicatively coupled gate, up, and down projections of each SwiGLU expert to estimate its transformation capacity without calibration data (Sec.~\ref{sec:gsp}). Second, Router-Conditioned Refinement (RCR) constructs expert-specific directional prototypes from centered router weights and evaluates experts along their routing-preferred directions, compensating for directional specialization missed by GSP (Sec.~\ref{sec:rcr}). Finally, ACE combines runtime router gates with offline GSP and RCR scores, skipping an expert slot only when both views identify it as low-contribution. 
Detailed pseudocode is provided in Appendix Sec.~\ref{sec:Algorithmic}.

\subsection{Global Spectral Proxy (GSP)} 
\label{sec:gsp} 
In standard MoE inference, router gates typically serve as the primary indicator of expert importance, determining a fixed top-$k$ activation pattern. However, a gate reflects only the relative \emph{routing preference} of the current token among candidate experts, rather than the intrinsic nonlinear transformation capacity of the expert networks. Thus, high routing affinity does not necessarily imply a large contribution to the MoE output. Gate-only skipping may therefore conflate routing preference with marginal contribution, while overlooking differences in the parameter structures and response capacities of individual experts.
To mitigate this measurement bias, ACE introduces a static transformation-capacity proxy derived exclusively from intrinsic expert parameters. Without modifying the original top-$k$ candidate set selected by the pretrained router, this proxy enables fine-grained, adaptive expert skipping. Specifically, for the $l$-th MoE layer and the $t$-th token, let $\{e_i\}_{i=1}^{k}$ denote the original top-$k$ experts selected by the pretrained router, with gate values $\{g_{l,t,i}\}_{i=1}^{k}$. Within this candidate set, we define the unnormalized GSP contribution score of the $i$-th routed expert slot as
\begin{equation}
s_{l,t,i}^{\mathrm{GSP}}
=
g_{l,t,i}\cdot A_l^{\mathrm{GSP}}(e_i),
\label{eq:gsp_contribution}
\end{equation}
where $g_{l,t,i}$ represents the token-dependent routing preference for expert $e_i$, while $A_l^{\mathrm{GSP}}(e_i)$ is a static amplification proxy derived solely from intrinsic model parameters. The latter characterizes the expert's transformation capacity for a normalized hidden representation. By combining dynamic routing preference with static transformation capacity, GSP distinguishes whether an expert is strongly preferred by the router from whether it is structurally capable of producing a strong response. Consequently, an expert slot receives a high GSP contribution score only when it exhibits both strong routing compatibility and high intrinsic transformation capacity.

To construct $A_l^{\mathrm{GSP}}(e)$ without calibration data, we analyze the parameterized input--output transformation of a SwiGLU expert. Since the RMSNorm preceding each MoE module maintains hidden states at a relatively stable scale, the expert input can be written as $x=\bar{x}\Gamma_l$, where $\bar{x}$ denotes the RMS-normalized input direction and $\Gamma_l$ is the diagonal matrix formed by the learned RMSNorm scaling vector at layer $l$. Consider a SwiGLU expert with gate, up, and down projection weights $W_{l,e}^{\mathrm{gate}}$, $W_{l,e}^{\mathrm{up}}$, and $W_{l,e}^{\mathrm{down}}$, respectively. Ignoring bias terms, its forward transformation is
\begin{equation}
f_{l,e}(x) = \left( \operatorname{SiLU} \left( \bar{x}\Gamma_l W_{l,e}^{\mathrm{gate}} \right) \odot \bar{x}\Gamma_l W_{l,e}^{\mathrm{up}} \right) W_{l,e}^{\mathrm{down}}.
\label{eq:swiglu_expert}
\end{equation}
Since $|\operatorname{SiLU}(z)| \leq |z|$ for any scalar $z$, the expert output admits the upper bound is
\begin{equation}
\left\|f_{l,e}(x)\right\|_2 \leq \left\|\bar{x}\right\|_2^2 \left\|\Gamma_l W_{l,e}^{\mathrm{gate}}\right\|_2 \left\|\Gamma_l W_{l,e}^{\mathrm{up}}\right\|_2 \left\|W_{l,e}^{\mathrm{down}}\right\|_2.
\label{eq:swiglu_upper_bound}
\end{equation}
This bound shows that the response capacity of an expert is jointly determined by the gate branch, up branch, and down projection through multiplicative coupling. Hence, the norm of any individual projection matrix is insufficient to characterize the transformation capacity of a SwiGLU expert. The complete derivation is provided in Appendix Sec.~\ref{app:swiglu_bound}.

Motivated by this multiplicative structure, GSP constructs a branch-symmetric matrix-norm surrogate requiring neither input activations nor calibration samples. Specifically, we estimate each expert's global response capacity through two complementary factorization paths:
\begin{align}
a_{l,e}^{\mathrm{up}} &= \left\|\Gamma_l W_{l,e}^{\mathrm{up}}\right\|_F \cdot \left\| \Gamma_l W_{l,e}^{\mathrm{gate}} W_{l,e}^{\mathrm{down}} \right\|_F,
\label{eq:gsp_up_path}
\\
a_{l,e}^{\mathrm{gate}} &= \left\|\Gamma_l W_{l,e}^{\mathrm{gate}}\right\|_F \cdot \left\| \Gamma_l W_{l,e}^{\mathrm{up}} W_{l,e}^{\mathrm{down}} \right\|_F.
\label{eq:gsp_gate_path}
\end{align}
Here, $a_{l,e}^{\mathrm{up}}$ combines the response scale of the up branch with the gate-to-down transmission path, while $a_{l,e}^{\mathrm{gate}}$ symmetrically combines the gate-branch scale with the up-to-down path. Together, these quantities provide complementary factorizations of the multiplicative SwiGLU transformation, preserving structural information from both input branches and the shared output projection. We aggregate them using the geometric mean:
\begin{equation}
\widetilde{a}_{l,e}^{\mathrm{GSP}} = \sqrt{a_{l,e}^{\mathrm{up}}a_{l,e}^{\mathrm{gate}}}.
\label{eq:gsp_geometric_mean}
\end{equation}
The geometric mean aligns naturally with SwiGLU's multiplicative structure and is more sensitive to imbalanced responses between the two complementary paths. Compared with additive aggregation, it limits the dominance of an exceptionally large single-path norm, assigning higher scores to experts with consistently strong responses along both paths. Since parameter scales, hidden dimensions, and RMSNorm scaling factors can vary substantially across MoE layers, raw surrogate values are not directly comparable. We therefore normalize $\widetilde{a}_{l,e}^{\mathrm{GSP}}$ by the layer-wise mean:
\begin{equation}
A_l^{\mathrm{GSP}}(e) =
\frac{\widetilde{a}_{l,e}^{\mathrm{GSP}}}
{\frac{1}{N_l}\sum_{j=1}^{N_l}\widetilde{a}_{l,j}^{\mathrm{GSP}}+\epsilon},
\label{eq:gsp_layer_normalization}
\end{equation}
where $N_l$ is the number of experts in the $l$-th MoE layer and $\epsilon$ ensures numerical stability. The normalized score $A_l^{\mathrm{GSP}}(e)$ measures the global structural response of expert $e$ relative to the layer average: $A_l^{\mathrm{GSP}}(e)>1$ indicates above-average static amplification capacity, whereas $A_l^{\mathrm{GSP}}(e)<1$ indicates a relatively weak global response.

During online inference, GSP requires neither additional expert forward passes nor activation-statistics collection. For each expert slot selected by the pretrained router, ACE retrieves the corresponding static amplification score $A_l^{\mathrm{GSP}}(e_i)$, precomputed once before deployment, and combines it with the current gate value. The resulting scores are normalized within the original top-$k$ candidate set to obtain the relative GSP contribution distribution:
\begin{equation}
p_{l,t,i}^{\mathrm{GSP}} =
\frac{s_{l,t,i}^{\mathrm{GSP}}}
{\sum_{j=1}^{k}s_{l,t,j}^{\mathrm{GSP}}+\epsilon}.
\label{eq:gsp_probability}
\end{equation}
Here, $p_{l,t,i}^{\mathrm{GSP}}$ measures the relative contribution of the $i$-th expert slot under the joint effect of token-dependent routing preference and global structural response capacity. Given a global threshold $\tau(q)$ for target skipping rate $q$, a slot satisfying $p_{l,t,i}^{\mathrm{GSP}}<\tau(q)$ is identified as a low-contribution candidate from the GSP perspective. Because all static scores are computed before deployment, online GSP requires only expert-indexed table lookup, scalar multiplication, and normalization over the original top-$k$ candidates. It therefore adds no high-dimensional expert computation and provides ACE with a calibration-free, parameter-intrinsic, and deployment-friendly baseline for global expert contribution estimation.

\subsection{Router-Conditioned Refinement (RCR)}
\label{sec:rcr}

Although GSP provides a calibration-free baseline for estimating global expert contributions, it is inherently \emph{unconditional}: it evaluates an expert's overall response to generic normalized input directions based on its parameter structure, without distinguishing the input subspaces that the expert receives. In practice, however, an MoE router often exhibits pronounced directional selectivity, assigning different experts to distinct regions of the hidden-state space. Consequently, an expert with a weak global structural response may not be uniformly weak across all directions; it may still produce a strong response along directions favored by the router. Relying solely on GSP may therefore underestimate experts with pronounced \emph{directional specialization}. To address this limitation, we introduce Router-Conditioned Refinement (RCR), which exploits directional preferences encoded by the pretrained router to restrict expert evaluation from the generic input space to the corresponding router-conditioned subspace, thereby providing a local directional correction to the global GSP estimate.

To extract expert-specific routing directions without accessing real activations, RCR directly leverages the parameter structure of the pretrained router. Consider the routing logit assigned to expert $e$ at layer $l$: $z_{l,e}(x)=x^{\top}w_{l,e}+b_{l,e}$, where $w_{l,e}$ denotes the router weight vector associated with expert $e$. Since softmax routing depends only on relative differences among expert logits, the router weights are invariant to a shared translation: adding the same vector to all expert weight vectors leaves the resulting routing probabilities unchanged. We therefore center the router weights within each layer to obtain a direction representation with explicit relative discriminative meaning:
\begin{equation}
\widetilde{w}_{l,e}
=
w_{l,e}
-
\frac{1}{N_l}
\sum_{j=1}^{N_l}w_{l,j}.
\label{eq:rcr_centered_router}
\end{equation}
The centered vector $\widetilde{w}_{l,e}$ characterizes the routing-discriminative direction of expert $e$ relative to other experts in the same layer, while remaining invariant to any shared translation of the router weights (proof is provided in Appendix Sec.~\ref{app:router_translation}). To interpret this centered direction from a conditional-distribution perspective, we adopt a Gaussian approximation to the centered hidden states. Let $\mu_l^x=\mathbb{E}_{x\sim p_l}[x]$ and assume $x-\mu_l^x\sim\mathcal{N}(0,\Sigma_l)$, where $p_l(x)$ denotes the input distribution at layer $l$ and $\Sigma_l$ is its covariance matrix. We then define an exponentially tilted distribution induced by the centered router direction:
\begin{equation}
p_{l,e}(x)
=
\frac{
\exp\left(\widetilde{w}_{l,e}^{\top}x\right)p_l(x)
}{
\mathbb{E}_{x\sim p_l}
\left[
\exp\left(\widetilde{w}_{l,e}^{\top}x\right)
\right]
}.
\label{eq:rcr_exponential_tilting}
\end{equation}
Under the Gaussian approximation, the conditional mean shift is $\mathbb{E}_{x\sim p_{l,e}}[x]-\mathbb{E}_{x\sim p_l}[x]=\Sigma_l\widetilde{w}_{l,e}$. When $\Sigma_l\approx\sigma_l^2I$, the mean-shift direction is approximately aligned with $\widetilde{w}_{l,e}$. Therefore, the centered router weight can be interpreted as a data-free approximation to the expert-conditioned input direction (complete derivation is provided in Appendix Sec.~\ref{app:gaussian_tilting}).

Based on the above analysis, RCR uses the centered router weight to approximate the expert-conditioned input direction and constructs an expert-specific direction prototype through RMS normalization:
\begin{equation}
q_{l,e}
=
\frac{
\widetilde{w}_{l,e}
}{
\operatorname{RMS}\!\left(\widetilde{w}_{l,e}\right)+\epsilon
}
\cdot\rho_l,
\label{eq:rcr_direction_prototype}
\end{equation}
where, for $w\in\mathbb{R}^{d}$, $\operatorname{RMS}(w)=\sqrt{\frac{1}{d}\sum_{j=1}^{d}w_j^2}$, $d$ denotes the hidden-state dimension, and $\rho_l$ is a prior on the input scale of layer $l$. To preserve the fully calibration-free property of ACE, we set $\rho_l=1$ throughout. RMS normalization removes variations in the overall magnitudes of router weight vectors across experts, allowing $q_{l,e}$ to primarily retain their relative routing directions rather than redundantly encode routing confidence, which is already captured by the runtime gate $g_{l,t,i}$. Since inputs following RMSNorm exhibit a relatively stable root-mean-square scale, $q_{l,e}$ serves as a data-free directional prototype whose scale is compatible with actual expert inputs. Before deployment, RCR feeds $q_{l,e}$ into its corresponding expert and performs a complete SwiGLU forward pass to measure the response along its routing-preferred direction. Specifically, the raw router-conditioned amplification of expert $e$ is defined as
\begin{equation}
\widetilde{a}_{l,e}^{\mathrm{RCR}}
=
\frac{
\left\|f_{l,e}\!\left(q_{l,e}\right)\right\|_2
}{
\left\|q_{l,e}\right\|_2+\epsilon
}.
\label{eq:rcr_raw_amplification}
\end{equation}
The numerator measures the output magnitude produced along the router-conditioned direction, while the denominator normalizes the prototype input scale. This ratio therefore characterizes the expert's directional amplification capacity and improves comparability across experts.

Similar to GSP, raw router-conditioned responses may vary across MoE layers due to differences in parameter magnitudes and nonlinear response ranges. We therefore normalize the prototype amplification within each layer:
\begin{equation}
A_l^{\mathrm{RCR}}(e)
=
\frac{
\widetilde{a}_{l,e}^{\mathrm{RCR}}
}{
\frac{1}{N_l}
\sum_{j=1}^{N_l}
\widetilde{a}_{l,j}^{\mathrm{RCR}}
+
\epsilon
}.
\label{eq:rcr_layer_normalization}
\end{equation}
The normalized score $A_l^{\mathrm{RCR}}(e)$ measures the router-conditioned response strength of expert $e$ relative to the average expert in the same layer. A value greater than one indicates an above-average response along its routing-preferred direction, even if its global structural response measured by GSP is weak. During online inference, RCR introduces no prototype evaluation or expert forward computation. ACE only retrieves the precomputed score $A_l^{\mathrm{RCR}}(e_i)$ for each routed expert slot and combines it with the current gate:
\begin{equation}
s_{l,t,i}^{\mathrm{RCR}}
=
g_{l,t,i}A_l^{\mathrm{RCR}}(e_i).
\label{eq:rcr_contribution}
\end{equation}
The resulting scores are normalized within the top-$k$ candidate set to obtain the router-conditioned contribution distribution:
\begin{equation}
p_{l,t,i}^{\mathrm{RCR}}
=
\frac{
s_{l,t,i}^{\mathrm{RCR}}
}{
\sum_{j=1}^{k}s_{l,t,j}^{\mathrm{RCR}}+\epsilon
}.
\label{eq:rcr_probability}
\end{equation}
Here, $p_{l,t,i}^{\mathrm{RCR}}$ measures the relative contribution of the $i$-th expert slot by considering token-dependent routing preference and the expert response along its router-conditioned direction.

GSP and RCR assess each expert slot from two complementary perspectives: GSP captures the expert's global transformation capacity encoded by its parameter structure, whereas RCR measures its localized response along the routing-preferred direction. To combine these estimates without additional fusion hyperparameters, ACE adopts a conservative maximum criterion and defines the final contribution score as
\begin{equation}
c_{l,t,i}^{\mathrm{ACE}}
=
\max\left(
p_{l,t,i}^{\mathrm{GSP}},
p_{l,t,i}^{\mathrm{RCR}}
\right).
\label{eq:ace_contribution}
\end{equation}
Under this criterion, a routed expert slot is considered skippable only when both GSP and RCR assign it low contribution. Given the global threshold $\tau(q)$ associated with target skipping rate $q$, slot $i$ is skipped only if $c_{l,t,i}^{\mathrm{ACE}}<\tau(q)$. This conservative aggregation prevents an expert from being removed when it is considered important from either the global structural or router-conditioned directional perspective. Equivalently, let the low-contribution candidate sets identified by GSP and RCR be
\begin{align}
\mathcal{S}_{l,t}^{\mathrm{GSP}}
&=
\left\{
i\;\middle|\;
p_{l,t,i}^{\mathrm{GSP}}<\tau(q)
\right\},
\label{eq:gsp_candidate_set}
\\
\mathcal{S}_{l,t}^{\mathrm{RCR}}
&=
\left\{
i\;\middle|\;
p_{l,t,i}^{\mathrm{RCR}}<\tau(q)
\right\}.
\label{eq:rcr_candidate_set}
\end{align}
According to Eq.~\eqref{eq:ace_contribution}, the final skipping set is
\begin{equation}
\mathcal{S}_{l,t}^{\mathrm{ACE}}
=
\mathcal{S}_{l,t}^{\mathrm{GSP}}
\cap
\mathcal{S}_{l,t}^{\mathrm{RCR}}.
\label{eq:ace_candidate_intersection}
\end{equation}
Thus, an expert slot is skipped only if both perspectives identify it as low-contribution. This intersection rule provides a conservative safeguard against incorrectly skipping direction-specialized experts whose global response may be weak but whose router-conditioned response remains strong (analysis of its conservativeness and resulting output perturbation is provided in Appendix Sec.~\ref{app:ace_conservative}). To preserve the routing structure of the pretrained model, ACE always retains the top-1 expert with the highest original router gate and optionally enforces a minimum number of active experts per token. After the skipping decision, the gate values of the remaining experts are renormalized before final MoE output aggregation. Through this complementary global--local verification, ACE reduces the risk of excessively skipping direction-specialized experts while enabling fine-grained, token-adaptive expert sparsification. It thereby preserves both global structural stability and router-conditioned directional fidelity without modifying the pretrained router or expert parameters.

%% file: Section/5_experiment.tex
\section{Experiments}
\label{sec:experiments}
We evaluate ACE along three dimensions: quality preservation at matched expert-skipping budgets, the contribution of its two parameter-derived views, and realized inference efficiency. The main text presents one complete 10--60\% budget table per model, component ablations, threshold-construction cost, and complete latency measurements. Appendix~\ref{app:experiments} contains all task-level results and threshold construction details.

\subsection{Experimental Setup}
\label{sec:experimental_setup}
\input{Tab/main_qwen3}
\paragraph{Models and benchmarks.}
We study Qwen3-30B-A3B-Instruct-2507, Qwen3.6-35B-A3B, and Gemma-4-26B-A4B-it~\citep{yang2025qwen3,qwen36modelcard,gemma4modelcard}. We evaluate WikiText-2 PPL at sequence length 2048~\citep{merity2016pointer} and 7 downstream tasks: ARC-Challenge (ARC-C), ARC-Easy (ARC-E), PIQA, MATH-500, GPQA-Diamond, HumanEval, and LiveCodeBench~\citep{clark2018arc,bisk2020piqa,hendrycks2021math,rein2024gpqa,chen2021codex,jain2024livecodebench}. We report unweighted mean of seven task accuracies. We omit Gemma-4 PPL because unusually high perplexity for the original model is poorly aligned with its downstream capability; consequently, downstream task accuracy provides a more informative measure of pruning-induced quality changes for this model.

\paragraph{Evaluation protocol.}
All methods use BF16 and the same Hugging Face and EvalScope 1.4.1 evaluation pipeline. All downstream evaluations are zero-shot, use identical prompts and splits, and apply greedy decoding. Generation is capped at 1,024 tokens for ARC-C, ARC-E, and PIQA and at 2,048 tokens for reasoning and coding tasks. Results are single deterministic runs with seed 42; sub-point differences, especially on GPQA-Diamond, should therefore be interpreted cautiously.
The skipping ratio is the fraction of routed top-$k$ expert slots that are not executed. For every method and benchmark, a single unlabeled full-model pass collects runtime selection scores and searches for the threshold realizing each operating point in $\{10,20,30,40,50,60\}\%$. Accordingly, the ratios reported in all result tables are measured, realized skipping ratios rather than unconstrained target values. The top-1 routed expert is always retained, and threshold search accounts for this safeguard and the minimum-active-expert constraint.
\input{Tab/main_qwen36}

ACE's contribution estimators remain entirely data-free: GSP and RCR are computed only from frozen model parameters and require neither training samples nor evaluation inputs. The unlabeled pass is used solely to translate a requested execution budget into the numerical scale of resulting scores; it never defines, trains, or updates either estimator. Moreover, thresholds associated with the same realized budget remain closely clustered across workloads and can therefore be transferred across datasets. Appendix~\ref{app:threshold_mapping} formalizes the mapping procedure and reports its cross-workload stability.

\paragraph{Baselines.}
We compare ACE with router-score skipping (Score), NAEE~\citep {lu2024not}, MoDES~\citep {huang2026modes}, and DiEP~\citep {bai2026diep}. On Qwen3-30B, where compatible runs are available, we additionally report AIMER~\citep {liu2026aimer}, Top-$P$ routing~\citep{huang2024harder}, SERE~\citep{wu2026sere}, XShare~\citep{vankov2026xshare}, and ExpertSparsity~\citep{lu2024not}. All methods preserve pretrained top-$k$ candidate set and are evaluated with the same BF16 implementation, realized skipping budgets, prompts, splits, and decoding pipeline. This controlled protocol ensures performance differences primarily reflect the quality of expert-selection decisions rather than differences in evaluation conditions.

\subsection{Main Results}
\label{sec:main_results}
Tables~\ref{tab:main_qwen3}--\ref{tab:main_gemma} report the complete 10--60\% curves for all three models. Under mild skipping, several methods remain close to the unpruned model because only the least influential routed slots are removed. For example, Score attains the highest average accuracy on Qwen3.6 at 10\%, while ACE remains competitive. This result indicates that router confidence can be sufficient in the low-sparsity regime and that ACE's structural estimates become most useful when increasingly consequential expert decisions must be made.

As the skipping ratio increases, the difference between router-only or single-view criteria and ACE becomes substantially larger. ACE achieves the best average accuracy on Qwen3-30B (tied with NAEE at 10\%) and from 20\% onward on Qwen3.6, where it also obtains the lowest PPL throughout. At 50\%, ACE improves Qwen3.6 from MoDES's 9.42/71.42 to 8.67/75.57 (PPL/Acc.). At 60\%, it leads the strongest external baseline by 1.07 and 1.66 accuracy points on Qwen3-30B and Qwen3.6, respectively. These widening margins support the central motivation of ACE: high-ratio skipping requires more than relative router preference because structurally weak and directionally specialized experts cannot be reliably distinguished by gate values alone.

\begingroup
\parshape=17
  0.66\columnwidth 0.34\columnwidth
  0.66\columnwidth 0.34\columnwidth
  0.66\columnwidth 0.34\columnwidth
  0.66\columnwidth 0.34\columnwidth
  0.66\columnwidth 0.34\columnwidth
  0.66\columnwidth 0.34\columnwidth
  0.66\columnwidth 0.34\columnwidth
  0.66\columnwidth 0.34\columnwidth
  0.66\columnwidth 0.34\columnwidth
  0.66\columnwidth 0.34\columnwidth
  0.66\columnwidth 0.34\columnwidth
  0.66\columnwidth 0.34\columnwidth
  0.66\columnwidth 0.34\columnwidth
  0.66\columnwidth 0.34\columnwidth
  0.66\columnwidth 0.34\columnwidth
  0.66\columnwidth 0.34\columnwidth
  0pt \columnwidth
\noindent
\raisebox{0pt}[0pt][0pt]{%
  \makebox[0pt][r]{%
    \begin{minipage}[t]{0.64\columnwidth}
    \captionsetup{type=table,skip=2pt}
    \captionof{table}{Gemma-4 average accuracy across 10--60\% skipping.}
    \label{tab:main_gemma}
    \input{Tab/main_gemma}
    \end{minipage}%
    \hspace{0.02\columnwidth}%
  }%
}%
\footnotesize
Table~\ref{tab:main_gemma} reports average downstream accuracy; BF16 is excluded from ranking and bold marks the best skipping result. ACE leads at every budget. Its advantage over GSP grows from 0.44 points at 30\% to 1.38 points at 50\% and 2.46 points at 60\%. Competing criteria degrade sharply beyond 40\%, whereas the conservative maximum rule preserves experts that remain important under either structural or router-conditioned evidence.
\par
\endgroup
\smallskip

The results also reveal that language-modeling perplexity and downstream accuracy are complementary rather than interchangeable. A method with competitive WikiText-2 PPL does not necessarily preserve reasoning and coding accuracy, particularly at aggressive budgets. ACE is consistently strong on both measures, suggesting that its conservative maximum fusion avoids removing experts that appear unimportant under only one contribution view. Complete task-level results supporting these trends are provided in Appendix~\ref{app:complete_task_results}.

\noindent
\begin{minipage}[t]{\columnwidth}
\subsection{Ablation Study}
\label{sec:ablation}

\begingroup
\parshape=18
  0.58\columnwidth 0.42\columnwidth
  0.58\columnwidth 0.42\columnwidth
  0.58\columnwidth 0.42\columnwidth
  0.58\columnwidth 0.42\columnwidth
  0.58\columnwidth 0.42\columnwidth
  0.58\columnwidth 0.42\columnwidth
  0.58\columnwidth 0.42\columnwidth
  0.58\columnwidth 0.42\columnwidth
  0.58\columnwidth 0.42\columnwidth
  0.58\columnwidth 0.42\columnwidth
  0.58\columnwidth 0.42\columnwidth
  0.58\columnwidth 0.42\columnwidth
  0.58\columnwidth 0.42\columnwidth
  0.58\columnwidth 0.42\columnwidth
  0.58\columnwidth 0.42\columnwidth
  0.58\columnwidth 0.42\columnwidth
  0.58\columnwidth 0.42\columnwidth
  0pt \columnwidth
\noindent
\raisebox{0pt}[0pt][0pt]{%
  \makebox[0pt][r]{%
    \begin{minipage}[t]{0.55\columnwidth}
    \captionsetup{type=table,skip=2pt}
    \captionof{table}{Component ablation. BF16: PPL 7.52 and Avg. 80.67; Drop is the Avg. decrease.}
    \label{tab:ablation}
    \input{Tab/ablation}
    \end{minipage}%
    \hspace{0.03\columnwidth}%
  }%
}%
\footnotesize
Table~\ref{tab:ablation} isolates the two offline estimators. GSP provides the stronger individual signal, whereas RCR alone degrades more rapidly because a single router-derived prototype cannot represent the complete conditional input distribution. Their complementary value emerges in ACE: at 50\% skipping, the combined rule reduces PPL from 8.99 to 8.85 and improves average accuracy from 74.10 to 74.30 over GSP.
\par
\endgroup
\vspace{2.5\baselineskip}
\end{minipage}
\par
\smallskip

\noindent
\begin{minipage}[t]{\columnwidth}
\begingroup
\parshape=14
  0.58\columnwidth 0.42\columnwidth
  0.58\columnwidth 0.42\columnwidth
  0.58\columnwidth 0.42\columnwidth
  0.58\columnwidth 0.42\columnwidth
  0.58\columnwidth 0.42\columnwidth
  0.58\columnwidth 0.42\columnwidth
  0.58\columnwidth 0.42\columnwidth
  0.58\columnwidth 0.42\columnwidth
  0.58\columnwidth 0.42\columnwidth
  0.58\columnwidth 0.42\columnwidth
  0.58\columnwidth 0.42\columnwidth
  0.58\columnwidth 0.42\columnwidth
  0.58\columnwidth 0.42\columnwidth
  0pt \columnwidth
\noindent
\raisebox{0pt}[0pt][0pt]{%
  \makebox[0pt][r]{%
    \begin{minipage}[t]{0.55\columnwidth}
    \captionsetup{type=table,skip=2pt}
    \captionof{table}{Fusion-strategy ablation on Qwen3-30B-A3B. Max is the ACE rule.}
    \label{tab:fusion_ablation}
    \input{Tab/fusion_ablation}
    \end{minipage}%
    \hspace{0.03\columnwidth}%
  }%
}%
\footnotesize
Table~\ref{tab:fusion_ablation} shows that Max performs best. It implements a conservative intersection: a slot is skipped only when both views consider it dispensable. Min is overly aggressive, while averaging can suppress a strong signal from one view. Max preserves complementary evidence without an additional fusion hyperparameter.
\par
\endgroup
\end{minipage}
\par
\smallskip

\noindent
\begin{minipage}[t]{\columnwidth}
\subsection{Efficiency Analysis}
\label{sec:efficiency}

\begingroup
\parshape=11
  0.58\columnwidth 0.42\columnwidth
  0.58\columnwidth 0.42\columnwidth
  0.58\columnwidth 0.42\columnwidth
  0.58\columnwidth 0.42\columnwidth
  0.58\columnwidth 0.42\columnwidth
  0.58\columnwidth 0.42\columnwidth
  0.58\columnwidth 0.42\columnwidth
  0.58\columnwidth 0.42\columnwidth
  0.58\columnwidth 0.42\columnwidth
  0.58\columnwidth 0.42\columnwidth
  0pt \columnwidth
\noindent
\raisebox{0pt}[0pt][0pt]{%
  \makebox[0pt][r]{%
    \begin{minipage}[t]{0.55\columnwidth}
    \captionsetup{type=table,skip=2pt}
    \captionof{table}{Threshold selection on Qwen3-30B at 10\% skipping.}
    \label{tab:threshold_time}
    \input{Tab/threshold_time}
    \end{minipage}%
    \hspace{0.03\columnwidth}%
  }%
}%
\footnotesize
ACE precomputes GSP and RCR once; online inference uses only expert-indexed lookups and top-$k$ scalar operations. Table~\ref{tab:threshold_time} shows that one quantile pass constructs a threshold in 1.8 minutes, versus 16.5 minutes for binary search and 18.9 minutes for MoDES frontier search. This 9.2--10.5$\times$ reduction concerns deployment preparation rather than token-level inference.
\par
\endgroup
\end{minipage}
\par
\smallskip

\noindent
\begin{minipage}[t]{\columnwidth}
\begingroup
\parshape=12
  0.60\columnwidth 0.40\columnwidth
  0.60\columnwidth 0.40\columnwidth
  0.60\columnwidth 0.40\columnwidth
  0.60\columnwidth 0.40\columnwidth
  0.60\columnwidth 0.40\columnwidth
  0.60\columnwidth 0.40\columnwidth
  0.60\columnwidth 0.40\columnwidth
  0.60\columnwidth 0.40\columnwidth
  0.60\columnwidth 0.40\columnwidth
  0.60\columnwidth 0.40\columnwidth
  0.60\columnwidth 0.40\columnwidth
  0pt \columnwidth
\noindent
\raisebox{0pt}[0pt][0pt]{%
  \makebox[0pt][r]{%
    \begin{minipage}[t]{0.57\columnwidth}
    \captionsetup{type=table,skip=2pt}
    \captionof{table}{Latency at 60\% skipping.}
    \label{tab:latency}
    \input{Tab/latency}
    \end{minipage}%
    \hspace{0.03\columnwidth}%
  }%
}%
\footnotesize
On an NVIDIA A100, ACE provides 1.72--2.25$\times$ TTFT and 1.31--1.41$\times$ TPOT speedups. At length 1,024 and batch size 1, TTFT falls from 270.9 to 120.3 ms and TPOT from 85.1 to 63.0 ms/token. Matched-budget methods share the same dispatch path; lightweight scoring keeps their latency within 1\%. Thus, ACE improves quality at the same executed-expert cost.
\par
\endgroup
\end{minipage}
\par

%% file: Tab/main_qwen3.tex
\begin{table}[b]
\caption{Complete 10--60\% budget comparison on Qwen3-30B-A3B-Instruct-2507. W2 denotes WikiText-2 PPL and Acc. denotes average downstream accuracy. BF16 is excluded from ranking; bold marks the best skipping result per metric and budget.}
\label{tab:main_qwen3}
\centering
\resizebox{\columnwidth}{!}{%
\begin{tabular}{l*{6}{rr}}
\toprule
\multirow{2}{*}{Method} & \multicolumn{2}{c}{10\%} & \multicolumn{2}{c}{20\%} & \multicolumn{2}{c}{30\%} & \multicolumn{2}{c}{40\%} & \multicolumn{2}{c}{50\%} & \multicolumn{2}{c}{60\%} \\
\cmidrule(lr){2-3}\cmidrule(lr){4-5}\cmidrule(lr){6-7}\cmidrule(lr){8-9}\cmidrule(lr){10-11}\cmidrule(lr){12-13}
& W2$\downarrow$ & Acc.$\uparrow$ & W2$\downarrow$ & Acc.$\uparrow$ & W2$\downarrow$ & Acc.$\uparrow$ & W2$\downarrow$ & Acc.$\uparrow$ & W2$\downarrow$ & Acc.$\uparrow$ & W2$\downarrow$ & Acc.$\uparrow$ \\
\midrule
BF16 & 7.52 & 80.67 & 7.52 & 80.67 & 7.52 & 80.67 & 7.52 & 80.67 & 7.52 & 80.67 & 7.52 & 80.67 \\
\midrule
Score & 7.70 & 79.69 & 8.09 & 77.85 & 8.65 & 77.97 & 9.63 & 75.39 & 12.03 & 66.69 & 18.76 & 37.64 \\
NAEE & 7.70 & \textbf{80.39} & 8.07 & 79.34 & 8.68 & 77.77 & 9.82 & 72.93 & 12.48 & 64.40 & 19.56 & 43.44 \\
MoDES & 7.57 & 79.79 & 7.84 & 78.80 & 8.33 & 77.99 & 9.28 & 76.03 & 12.25 & 56.61 & 17.80 & 39.15 \\
DiEP & 10.09 & 76.58 & 16.00 & 64.36 & 28.42 & 47.05 & 65.87 & 27.44 & 136.92 & 3.96 & 459.88 & 3.62 \\
AIMER & 7.55 & 80.08 & \textbf{7.65} & 79.94 & \textbf{7.82} & 79.36 & 8.24 & 77.17 & 8.86 & 73.40 & 10.89 & 60.88 \\
Top-$P$ & \textbf{7.54} & 79.95 & 7.66 & 79.54 & \textbf{7.82} & 79.27 & \textbf{8.19} & 77.73 & \textbf{8.85} & 74.17 & 10.87 & 61.91 \\
SERE & 8.77 & 75.59 & 11.00 & 66.65 & 14.27 & 62.55 & 20.81 & 58.05 & 46.36 & 38.62 & 178.96 & 9.82 \\
XShare & 8.10 & 76.24 & 9.14 & 63.27 & 10.82 & 32.96 & 13.67 & 22.48 & 19.63 & 10.98 & 39.22 & 6.15 \\
ExpertSparsity & 7.68 & 79.54 & 8.02 & 79.45 & 8.49 & 78.63 & 9.31 & 76.76 & 10.98 & 73.67 & 13.39 & 62.28 \\
GSP & 7.56 & 80.14 & 7.67 & 79.94 & 7.87 & 79.59 & 8.22 & 77.86 & 8.99 & 74.10 & 10.99 & 62.53 \\
RCR & 7.59 & 80.13 & 7.83 & 78.85 & 8.31 & 77.81 & 9.19 & 75.36 & 11.62 & 67.01 & 17.95 & 39.60 \\
ACE & 7.55 & \textbf{80.39} & 7.66 & \textbf{80.25} & 7.86 & \textbf{79.77} & \textbf{8.19} & \textbf{78.21} & \textbf{8.85} & \textbf{74.30} & \textbf{10.86} & \textbf{63.35} \\
\bottomrule
\end{tabular}%
}
\end{table}

%% file: Tab/main_qwen36.tex
\begin{table}[b]
\caption{Complete 10--60\% budget comparison on Qwen3.6-35B-A3B. W2 denotes WikiText-2 PPL and Acc. denotes average downstream accuracy. BF16 is excluded from ranking; bold marks the best skipping result per metric and budget.}
\label{tab:main_qwen36}
\centering
\resizebox{\columnwidth}{!}{%
\begin{tabular}{l*{6}{rr}}
\toprule
\multirow{2}{*}{Method} & \multicolumn{2}{c}{10\%} & \multicolumn{2}{c}{20\%} & \multicolumn{2}{c}{30\%} & \multicolumn{2}{c}{40\%} & \multicolumn{2}{c}{50\%} & \multicolumn{2}{c}{60\%} \\
\cmidrule(lr){2-3}\cmidrule(lr){4-5}\cmidrule(lr){6-7}\cmidrule(lr){8-9}\cmidrule(lr){10-11}\cmidrule(lr){12-13}
& W2$\downarrow$ & Acc.$\uparrow$ & W2$\downarrow$ & Acc.$\uparrow$ & W2$\downarrow$ & Acc.$\uparrow$ & W2$\downarrow$ & Acc.$\uparrow$ & W2$\downarrow$ & Acc.$\uparrow$ & W2$\downarrow$ & Acc.$\uparrow$ \\
\midrule
BF16 & 7.01 & 81.17 & 7.01 & 81.17 & 7.01 & 81.17 & 7.01 & 81.17 & 7.01 & 81.17 & 7.01 & 81.17 \\
\midrule
Score & 7.38 & \textbf{81.66} & 7.79 & 80.51 & 8.36 & 78.41 & 9.21 & 74.61 & 10.58 & 67.45 & 13.38 & 54.66 \\
NAEE & 7.37 & 81.27 & 7.80 & 80.99 & 8.40 & 79.01 & 9.30 & 74.84 & 10.89 & 67.50 & 14.17 & 53.42 \\
MoDES & 7.38 & 80.86 & 7.41 & 80.59 & 7.72 & 78.19 & 8.10 & 75.74 & 9.42 & 71.42 & 10.87 & 69.04 \\
DiEP & 8.83 & 79.18 & 12.09 & 76.91 & 17.62 & 72.73 & 28.45 & 66.16 & 45.43 & 56.42 & 85.12 & 32.27 \\
GSP & 7.18 & 81.12 & 7.38 & 80.50 & 7.66 & 79.46 & 8.08 & 78.25 & 8.87 & 75.01 & 10.48 & 70.54 \\
RCR & 7.31 & 80.97 & 7.65 & 81.05 & 8.17 & 78.36 & 8.95 & 75.35 & 10.39 & 66.42 & 13.48 & 45.38 \\
ACE & \textbf{7.16} & 81.41 & \textbf{7.34} & \textbf{81.46} & \textbf{7.57} & \textbf{79.89} & \textbf{7.88} & \textbf{78.79} & \textbf{8.67} & \textbf{75.57} & \textbf{9.98} & \textbf{70.70} \\
\bottomrule
\end{tabular}%
}
\end{table}

%% file: Tab/main_gemma.tex
\centering
\scriptsize
\setlength{\tabcolsep}{1.8pt}
\renewcommand{\arraystretch}{0.92}
\resizebox{\linewidth}{!}{%
\begin{tabular}{lcccccc}
\toprule
Method & \multicolumn{6}{c}{Acc.$\uparrow$ (\%)} \\
\cmidrule(lr){2-7}
& 10\% & 20\% & 30\% & 40\% & 50\% & 60\% \\
\midrule
BF16 & 81.63 & 81.63 & 81.63 & 81.63 & 81.63 & 81.63 \\
\midrule
Score & 80.59 & 80.14 & 78.19 & 76.10 & 71.14 & 64.05 \\
NAEE & 80.67 & 80.73 & 77.97 & 73.84 & 68.28 & 57.36 \\
MoDES & 80.82 & 80.63 & 78.42 & 77.84 & 38.74 & 26.86 \\
DiEP & 79.61 & 78.66 & 75.90 & 72.14 & 30.83 & 18.13 \\
GSP & 81.14 & 81.16 & 79.71 & 78.65 & 75.73 & 69.65 \\
RCR & 80.74 & 79.84 & 78.00 & 77.49 & 68.96 & 60.03 \\
ACE & \textbf{81.59} & \textbf{81.28} & \textbf{80.15} & \textbf{79.51} & \textbf{77.11} & \textbf{72.11} \\
\bottomrule
\end{tabular}
}

%% file: Tab/ablation.tex
\centering
\scriptsize
\setlength{\tabcolsep}{2.0pt}
\renewcommand{\arraystretch}{0.92}
\resizebox{\linewidth}{!}{%
\begin{tabular}{ccccc}
\toprule
Skip & Method & PPL$\downarrow$ & Avg.$\uparrow$ & Drop \\
\midrule
\multirow{3}{*}{10\%}
& GSP & 7.56 & 80.14 & $0.53$ \\
& RCR & 7.59 & 80.13 & $0.54$ \\
& ACE & \textbf{7.55} & \textbf{80.39} & $\mathbf{0.28}$ \\
\midrule
\multirow{3}{*}{30\%}
& GSP & 7.87 & 79.59 & $1.08$ \\
& RCR & 8.31 & 77.81 & $2.86$ \\
& ACE & \textbf{7.86} & \textbf{79.77} & $\mathbf{0.90}$ \\
\midrule
\multirow{3}{*}{50\%}
& GSP & 8.99 & 74.10 & $6.57$ \\
& RCR & 11.62 & 67.01 & $13.66$ \\
& ACE & \textbf{8.85} & \textbf{74.30} & $\mathbf{6.37}$ \\
\bottomrule
\end{tabular}
}

%% file: Tab/fusion_ablation.tex
\centering
\scriptsize
\setlength{\tabcolsep}{1.8pt}
\renewcommand{\arraystretch}{0.92}
\resizebox{\linewidth}{!}{%
\begin{tabular}{lrrrr}
\toprule
\multirow{2}{*}{Fusion} & \multicolumn{2}{c}{30\%} & \multicolumn{2}{c}{50\%} \\
\cmidrule(lr){2-3}\cmidrule(lr){4-5}
& PPL$\downarrow$ & Avg.$\uparrow$ & PPL$\downarrow$ & Avg.$\uparrow$ \\
\midrule
Min & 8.21 & 77.20 & 9.20 & 73.12 \\
Mean & 7.95 & 78.45 & 8.97 & 74.01 \\
Max (ACE) & \textbf{7.86} & \textbf{79.77} & \textbf{8.85} & \textbf{74.30} \\
\bottomrule
\end{tabular}
}

%% file: Tab/threshold_time.tex
\centering
\scriptsize
\setlength{\tabcolsep}{2.0pt}
\renewcommand{\arraystretch}{0.92}
\resizebox{\linewidth}{!}{%
\begin{tabular}{lrr}
\toprule
Procedure & Time (min) & Slowdown \\
\midrule
Quantile mapping & \textbf{1.8} & $\mathbf{1.0\times}$ \\
Binary search & 16.5 & $9.2\times$ \\
MoDES frontier search & 18.9 & $10.5\times$ \\
\bottomrule
\end{tabular}
}

%% file: Tab/latency.tex
\centering
\scriptsize
\setlength{\tabcolsep}{1.4pt}
\renewcommand{\arraystretch}{0.88}
\resizebox{\linewidth}{!}{%
\begin{tabular}{rrrrrr}
\toprule
\multirow{2}{*}{Length} & \multirow{2}{*}{Batch} & \multicolumn{2}{c}{TTFT (ms)} & \multicolumn{2}{c}{TPOT (ms/tok)} \\
\cmidrule(lr){3-4}\cmidrule(lr){5-6}
& & BF16 & ACE & BF16 & ACE \\
\midrule
256 & 1 & 144.7 & \textbf{84.1} & 34.9 & \textbf{26.7} \\
512 & 1 & 183.7 & \textbf{102.3} & 57.2 & \textbf{44.9} \\
1024 & 1 & 270.9 & \textbf{120.3} & 85.1 & \textbf{63.0} \\
1024 & 2 & 290.5 & \textbf{156.7} & 95.3 & \textbf{67.8} \\
1024 & 4 & 344.9 & \textbf{182.6} & 127.3 & \textbf{91.1} \\
\bottomrule
\end{tabular}
}

%% file: Section/6_conclusion.tex
\section{Conclusion}
\label{sec:conclusion}
We introduced ACE, a training-free and checkpoint-preserving method for token-adaptive expert skipping in MoE language models. GSP captures global SwiGLU transformation capacity, while RCR measures router-conditioned directional response; their conservative fusion skips a slot only when both views find it low-contribution. Across three MoE models, ACE is particularly effective at aggressive budgets. At 50\% skipping, it achieves the best average accuracy on all three models, reduces Qwen3.6 WikiText-2 perplexity by 7.96\% versus the strongest competitor, and accelerates prefill and decoding. Current results use workload-matched threshold mapping and single deterministic runs; future work should examine threshold transfer, repeated-run uncertainty, and distributed expert dispatch.

%% file: Section/07_appendix.tex
\section*{Appendix Overview}
\begin{itemize}
    \item 
\end{itemize}

\section{Appendix}
\label{sec:appendix}

\subsection{Detailed Algorithms of ACE}
\label{sec:Algorithmic}

This subsection presents the complete algorithmic workflow of ACE. We provide four pseudocode blocks corresponding to the overall framework, Global Spectral Proxy (GSP), Router-Conditioned Refinement (RCR), and online conservative expert skipping. All expert-level statistics used by GSP and RCR are computed once before deployment. During inference, ACE only performs expert-indexed table lookup and lightweight scalar operations before selectively executing the retained experts.

\subsubsection{Overall Workflow of ACE}
\label{app:overall_ace}

\paragraph{Pipeline description.}
ACE consists of an offline parameter-analysis stage and an online token-adaptive skipping stage. Given a pretrained MoE-based LLM, the offline stage first applies GSP to estimate the global transformation capacity of each expert from its frozen SwiGLU parameters. RCR then constructs expert-specific direction prototypes from the centered router weights and evaluates the corresponding experts along their routing-preferred directions. The resulting GSP and RCR amplification scores are stored as two expert-indexed lookup tables. During inference, ACE combines these offline scores with runtime router gates and skips an expert slot only when both perspectives identify it as low-contribution. The overall workflow is summarized in Algorithm~\ref{alg:ace_overall}.

\begin{algorithm}[t!]
\caption{ACE: Overall Calibration-Free Expert Skipping Framework}
\label{alg:ace_overall}
\KwIn{Pretrained MoE-based LLM $M$; target skipping control $q$; threshold mapping $\tau(\cdot)$; minimum number of active experts $m_{\min}$; numerical constant $\epsilon$}
\KwOut{GSP table $\mathcal{A}^{\mathrm{GSP}}$; RCR table $\mathcal{A}^{\mathrm{RCR}}$; ACE inference procedure $\widehat{M}$}

\tcp{Offline parameter analysis}
$\mathcal{A}^{\mathrm{GSP}}
\leftarrow
\mathrm{GSP}(M,\epsilon)$\;

$\mathcal{A}^{\mathrm{RCR}}
\leftarrow
\mathrm{RCR}(M,\epsilon)$\;

Store $\mathcal{A}^{\mathrm{GSP}}$ and
$\mathcal{A}^{\mathrm{RCR}}$ as expert-indexed lookup tables\;

\tcp{Online token-adaptive inference}
$\tau_q \leftarrow \tau(q)$\;

\For{each input sequence}{
    \For{each token $t$}{
        \For{each MoE layer $l$}{
            Obtain the original top-$k$ routed experts and gates
            $\{(e_i,g_{l,t,i})\}_{i=1}^{k}$\;

            $\mathcal{I}_{l,t},
            \{\widehat{g}_{l,t,i}\}_{i\in\mathcal{I}_{l,t}}
            \leftarrow
            \mathrm{ACESkip}
            \left(
            l,t,
            \{(e_i,g_{l,t,i})\}_{i=1}^{k},
            \mathcal{A}^{\mathrm{GSP}},
            \mathcal{A}^{\mathrm{RCR}},
            \tau_q,
            m_{\min},
            \epsilon
            \right)$\;

            Evaluate only the experts indexed by
            $\mathcal{I}_{l,t}$\;

            Aggregate their outputs using
            $\{\widehat{g}_{l,t,i}\}_{i\in\mathcal{I}_{l,t}}$\;
        }
    }
}

Construct the resulting inference procedure as $\widehat{M}$\;
\end{algorithm}

\subsubsection{Global Spectral Proxy}
\label{app:gsp_algorithm}

\paragraph{Global structural response estimation.}
GSP estimates each expert's intrinsic transformation capacity directly from its frozen parameters. For the $l$-th MoE layer, let $\Gamma_l$ denote the diagonal matrix formed by the learned RMSNorm scaling vector. For each expert $e$, GSP constructs two branch-symmetric surrogates that jointly preserve the gate, up, and down projections. Their geometric mean provides the raw global amplification score, which is subsequently normalized by the average score of all experts in the same layer.

\paragraph{Offline computation.}
The entire procedure requires neither calibration samples nor input activations. Once the normalized scores $A_l^{\mathrm{GSP}}(e)$ have been computed for all experts, they are stored as a static lookup table and reused for every token during inference. The complete procedure is given in Algorithm~\ref{alg:gsp}.

\begin{algorithm}[t!]
\caption{GSP: Global Spectral Proxy}
\label{alg:gsp}
\KwIn{Pretrained MoE-based LLM $M$ with $L$ MoE layers; numerical constant $\epsilon$}
\KwOut{Global amplification table
$\mathcal{A}^{\mathrm{GSP}}
=
\{A_l^{\mathrm{GSP}}(e)\}$}

Initialize
$\mathcal{A}^{\mathrm{GSP}}\leftarrow\varnothing$\;

\For{each MoE layer $l=1,\ldots,L$}{
    Extract the RMSNorm scaling vector $\gamma_l$\;

    $\Gamma_l\leftarrow\operatorname{Diag}(\gamma_l)$\;

    Let $N_l$ denote the number of experts in layer $l$\;

    \For{each expert $e=1,\ldots,N_l$}{
        Extract
        $W_{l,e}^{\mathrm{gate}}$,
        $W_{l,e}^{\mathrm{up}}$, and
        $W_{l,e}^{\mathrm{down}}$\;

        $\mathbf{G}_{l,e}
        \leftarrow
        \Gamma_l W_{l,e}^{\mathrm{gate}}$\;

        $\mathbf{U}_{l,e}
        \leftarrow
        \Gamma_l W_{l,e}^{\mathrm{up}}$\;

        $a_{l,e}^{\mathrm{up}}
        \leftarrow
        \left\|\mathbf{U}_{l,e}\right\|_F
        \cdot
        \left\|
        \mathbf{G}_{l,e}
        W_{l,e}^{\mathrm{down}}
        \right\|_F$\;

        $a_{l,e}^{\mathrm{gate}}
        \leftarrow
        \left\|\mathbf{G}_{l,e}\right\|_F
        \cdot
        \left\|
        \mathbf{U}_{l,e}
        W_{l,e}^{\mathrm{down}}
        \right\|_F$\;

        $\widetilde{a}_{l,e}^{\mathrm{GSP}}
        \leftarrow
        \sqrt{
        a_{l,e}^{\mathrm{up}}
        a_{l,e}^{\mathrm{gate}}
        }$\;
    }

    $\mu_l^{\mathrm{GSP}}
    \leftarrow
    \frac{1}{N_l}
    \sum_{j=1}^{N_l}
    \widetilde{a}_{l,j}^{\mathrm{GSP}}$\;

    \For{each expert $e=1,\ldots,N_l$}{
        $A_l^{\mathrm{GSP}}(e)
        \leftarrow
        \frac{
        \widetilde{a}_{l,e}^{\mathrm{GSP}}
        }{
        \mu_l^{\mathrm{GSP}}+\epsilon
        }$\;

        Add $A_l^{\mathrm{GSP}}(e)$ to
        $\mathcal{A}^{\mathrm{GSP}}$\;
    }
}

\Return{$\mathcal{A}^{\mathrm{GSP}}$}
\end{algorithm}

\subsubsection{Router-Conditioned Refinement}
\label{app:rcr_algorithm}

\paragraph{Router-conditioned direction construction.}
RCR refines the unconditional GSP estimate by measuring each expert along the direction preferred by the pretrained router. For the $l$-th layer, the router weight associated with expert $e$ is centered by subtracting the mean router weight over all experts. This removes the shared translation component and preserves the relative routing-discriminative direction. RCR then applies RMS normalization to obtain the expert-specific direction prototype $q_{l,e}$.

\paragraph{Directional response evaluation.}
Each prototype is passed once through its corresponding frozen SwiGLU expert before deployment. The ratio between the expert output norm and prototype input norm defines the raw router-conditioned amplification. As in GSP, the raw responses are normalized within each layer and stored as an expert-indexed lookup table. This offline computation is summarized in Algorithm~\ref{alg:rcr}.

\begin{algorithm}[t!]
\caption{RCR: Router-Conditioned Refinement}
\label{alg:rcr}
\KwIn{Pretrained MoE-based LLM $M$ with $L$ MoE layers; numerical constant $\epsilon$}
\KwOut{Router-conditioned amplification table
$\mathcal{A}^{\mathrm{RCR}}
=
\{A_l^{\mathrm{RCR}}(e)\}$}

Initialize
$\mathcal{A}^{\mathrm{RCR}}\leftarrow\varnothing$\;

\For{each MoE layer $l=1,\ldots,L$}{
    Let $N_l$ denote the number of experts in layer $l$\;

    Let $d$ denote the router input dimension\;

    Extract router weight vectors
    $\{w_{l,e}\}_{e=1}^{N_l}$\;

    $\overline{w}_l
    \leftarrow
    \frac{1}{N_l}
    \sum_{j=1}^{N_l}w_{l,j}$\;

    Set the calibration-free prototype scale
    $\rho_l\leftarrow1$\;

    \For{each expert $e=1,\ldots,N_l$}{
        $\widetilde{w}_{l,e}
        \leftarrow
        w_{l,e}-\overline{w}_l$\;

        $r_{l,e}
        \leftarrow
        \sqrt{
        \frac{1}{d}
        \sum_{u=1}^{d}
        \left(\widetilde{w}_{l,e,u}\right)^2
        }$\;

        $q_{l,e}
        \leftarrow
        \rho_l
        \frac{
        \widetilde{w}_{l,e}
        }{
        r_{l,e}+\epsilon
        }$\;

        $y_{l,e}^{\mathrm{proto}}
        \leftarrow
        f_{l,e}(q_{l,e})$\;

        $\widetilde{a}_{l,e}^{\mathrm{RCR}}
        \leftarrow
        \frac{
        \left\|y_{l,e}^{\mathrm{proto}}\right\|_2
        }{
        \left\|q_{l,e}\right\|_2+\epsilon
        }$\;
    }

    $\mu_l^{\mathrm{RCR}}
    \leftarrow
    \frac{1}{N_l}
    \sum_{j=1}^{N_l}
    \widetilde{a}_{l,j}^{\mathrm{RCR}}$\;

    \For{each expert $e=1,\ldots,N_l$}{
        $A_l^{\mathrm{RCR}}(e)
        \leftarrow
        \frac{
        \widetilde{a}_{l,e}^{\mathrm{RCR}}
        }{
        \mu_l^{\mathrm{RCR}}+\epsilon
        }$\;

        Add $A_l^{\mathrm{RCR}}(e)$ to
        $\mathcal{A}^{\mathrm{RCR}}$\;
    }
}

\Return{$\mathcal{A}^{\mathrm{RCR}}$}
\end{algorithm}

\paragraph{Calibration-free prototype scale.}
ACE sets $\rho_l=1$ for every MoE layer. This preserves the fully calibration-free setting and prevents the prototype magnitude from encoding unavailable activation statistics. Because each prototype is RMS-normalized, it retains the relative router direction while maintaining a stable input scale.

\subsubsection{Online Conservative Expert Skipping}
\label{app:online_skipping}

\paragraph{Dual-view contribution estimation.}
During inference, ACE retains the original top-$k$ candidate set selected by the pretrained router. For each routed expert slot, it combines the current router gate with the offline GSP and RCR amplification scores. The two resulting score sets are independently normalized within the top-$k$ candidate set, producing $p_{l,t,i}^{\mathrm{GSP}}$ and $p_{l,t,i}^{\mathrm{RCR}}$.

\paragraph{Conservative skipping rule.}
ACE defines the final contribution score as the maximum of the two normalized probabilities. Consequently, an expert slot is selected for skipping only when both views assign it a contribution below $\tau_q$. The original top-1 expert is always retained. If the threshold rule would leave fewer than $m_{\min}$ experts active, ACE restores the skipped candidates with the largest contribution scores until the minimum active-expert constraint is satisfied. Finally, the gates of the retained experts are renormalized before expert execution and output aggregation. Algorithm~\ref{alg:ace_skip} details this online procedure.

\begin{algorithm}[t!]
\caption{ACE: Online Conservative Expert Skipping}
\label{alg:ace_skip}
\KwIn{Layer index $l$; token index $t$; routed experts and gates $\{(e_i,g_{l,t,i})\}_{i=1}^{k}$; lookup tables $\mathcal{A}^{\mathrm{GSP}}$ and $\mathcal{A}^{\mathrm{RCR}}$; threshold $\tau_q$; minimum number of active experts $m_{\min}$; numerical constant $\epsilon$}
\KwOut{Active expert-slot set $\mathcal{I}_{l,t}$; renormalized gates $\{\widehat{g}_{l,t,i}\}_{i\in\mathcal{I}_{l,t}}$}

\For{each routed expert slot $i=1,\ldots,k$}{
    $s_{l,t,i}^{\mathrm{GSP}}
    \leftarrow
    g_{l,t,i}
    A_l^{\mathrm{GSP}}(e_i)$\;

    $s_{l,t,i}^{\mathrm{RCR}}
    \leftarrow
    g_{l,t,i}
    A_l^{\mathrm{RCR}}(e_i)$\;
}

$Z_{l,t}^{\mathrm{GSP}}
\leftarrow
\sum_{j=1}^{k}
s_{l,t,j}^{\mathrm{GSP}}+\epsilon$\;

$Z_{l,t}^{\mathrm{RCR}}
\leftarrow
\sum_{j=1}^{k}
s_{l,t,j}^{\mathrm{RCR}}+\epsilon$\;

\For{each routed expert slot $i=1,\ldots,k$}{
    $p_{l,t,i}^{\mathrm{GSP}}
    \leftarrow
    \frac{
    s_{l,t,i}^{\mathrm{GSP}}
    }{
    Z_{l,t}^{\mathrm{GSP}}
    }$\;

    $p_{l,t,i}^{\mathrm{RCR}}
    \leftarrow
    \frac{
    s_{l,t,i}^{\mathrm{RCR}}
    }{
    Z_{l,t}^{\mathrm{RCR}}
    }$\;

    $c_{l,t,i}^{\mathrm{ACE}}
    \leftarrow
    \max
    \left(
    p_{l,t,i}^{\mathrm{GSP}},
    p_{l,t,i}^{\mathrm{RCR}}
    \right)$\;
}

$\mathcal{S}_{l,t}^{\mathrm{GSP}}
\leftarrow
\left\{
i
\;\middle|\;
p_{l,t,i}^{\mathrm{GSP}}<\tau_q
\right\}$\;

$\mathcal{S}_{l,t}^{\mathrm{RCR}}
\leftarrow
\left\{
i
\;\middle|\;
p_{l,t,i}^{\mathrm{RCR}}<\tau_q
\right\}$\;

$\mathcal{S}_{l,t}^{\mathrm{ACE}}
\leftarrow
\mathcal{S}_{l,t}^{\mathrm{GSP}}
\cap
\mathcal{S}_{l,t}^{\mathrm{RCR}}$\;

$i_{\mathrm{top1}}
\leftarrow
\arg\max_{i\in\{1,\ldots,k\}}
g_{l,t,i}$\;

$\mathcal{S}_{l,t}^{\mathrm{ACE}}
\leftarrow
\mathcal{S}_{l,t}^{\mathrm{ACE}}
\setminus
\{i_{\mathrm{top1}}\}$\;

$\mathcal{I}_{l,t}
\leftarrow
\{1,\ldots,k\}
\setminus
\mathcal{S}_{l,t}^{\mathrm{ACE}}$\;

\If{$|\mathcal{I}_{l,t}|<m_{\min}$}{
    Sort the indices in
    $\mathcal{S}_{l,t}^{\mathrm{ACE}}$
    by $c_{l,t,i}^{\mathrm{ACE}}$ in descending order\;

    Restore the first
    $m_{\min}-|\mathcal{I}_{l,t}|$
    indices to $\mathcal{I}_{l,t}$\;
}

$Z_{l,t}^{\mathrm{gate}}
\leftarrow
\sum_{j\in\mathcal{I}_{l,t}}
g_{l,t,j}+\epsilon$\;

\For{each retained expert slot $i\in\mathcal{I}_{l,t}$}{
    $\widehat{g}_{l,t,i}
    \leftarrow
    \frac{
    g_{l,t,i}
    }{
    Z_{l,t}^{\mathrm{gate}}
    }$\;
}

\Return{$\mathcal{I}_{l,t},
\{\widehat{g}_{l,t,i}\}_{i\in\mathcal{I}_{l,t}}$}
\end{algorithm}

\paragraph{Sparsified MoE output.}
Let $\mathcal{I}_{l,t}$ denote the final active expert-slot set returned by Algorithm~\ref{alg:ace_skip}. The sparsified MoE output is computed as
\begin{equation}
y_{l,t}^{\mathrm{ACE}}
=
\sum_{i\in\mathcal{I}_{l,t}}
\widehat{g}_{l,t,i}
f_{l,e_i}(x_{l,t}).
\label{eq:ace_sparse_output}
\end{equation}
Because the skipping decision is completed before expert execution, only experts in $\mathcal{I}_{l,t}$ require high-dimensional SwiGLU computation. The additional online operations introduced by ACE consist only of expert-indexed table lookup, scalar multiplication, normalization over the original top-$k$ candidates, and small-set comparisons. ACE therefore preserves the pretrained router and expert parameters while introducing negligible control overhead relative to expert forward computation.

\subsection{Detailed Proofs}
\label{sec:proof}

This subsection provides the theoretical results supporting the design of ACE. We first derive the output-response upper bound of a SwiGLU expert, which motivates the multiplicatively coupled construction of Global Spectral Proxy (GSP). We then prove the shared-translation invariance of router weights and derive the conditional mean shift induced by exponential tilting under a Gaussian approximation, providing the theoretical basis for Router-Conditioned Refinement (RCR). Finally, we establish the conservativeness of the dual-view skipping criterion and analyze the output perturbation introduced by expert skipping and gate renormalization.

\subsubsection{Output-Response Bound of a SwiGLU Expert}
\label{app:swiglu_bound}

We derive the output-response bound used to motivate the branch-symmetric construction of GSP. Consider the $e$-th SwiGLU expert in the $l$-th MoE layer. After the preceding RMSNorm, its input is represented as $x=\bar{x}\Gamma_l$, where $\bar{x}$ is the RMS-normalized hidden direction and $\Gamma_l$ is the diagonal matrix formed by the learned RMSNorm scaling vector. Ignoring bias terms, the expert output is

\begin{equation}
f_{l,e}(x)
=
\left(
\operatorname{SiLU}
\left(
\bar{x}\Gamma_lW_{l,e}^{\mathrm{gate}}
\right)
\odot
\bar{x}\Gamma_lW_{l,e}^{\mathrm{up}}
\right)
W_{l,e}^{\mathrm{down}}.
\label{eq:app_swiglu_forward}
\end{equation}

For compactness, define

\begin{equation}
\begin{aligned}
u
&=
\bar{x}\Gamma_lW_{l,e}^{\mathrm{gate}},
\\
v
&=
\bar{x}\Gamma_lW_{l,e}^{\mathrm{up}}.
\end{aligned}
\label{eq:app_swiglu_uv}
\end{equation}

For any scalar $z$, the SiLU activation satisfies

\begin{equation}
\left|
\operatorname{SiLU}(z)
\right|
=
\left|
z\sigma(z)
\right|
\leq
|z|,
\label{eq:app_silu_scalar_bound}
\end{equation}

because the sigmoid function satisfies $0<\sigma(z)<1$. Applying this inequality element-wise gives

\begin{equation}
\left\|
\operatorname{SiLU}(u)
\right\|_2
\leq
\|u\|_2.
\label{eq:app_silu_vector_bound}
\end{equation}

For any two vectors $a,b\in\mathbb{R}^{m}$, their Hadamard product satisfies

\begin{equation}
\begin{aligned}
\|a\odot b\|_2^2
&=
\sum_{j=1}^{m}a_j^2b_j^2
\\
&\leq
\left(
\sum_{j=1}^{m}a_j^2
\right)
\left(
\sum_{j=1}^{m}b_j^2
\right)
\\
&=
\|a\|_2^2\|b\|_2^2,
\end{aligned}
\label{eq:app_hadamard_bound}
\end{equation}

and therefore

\begin{equation}
\|a\odot b\|_2
\leq
\|a\|_2\|b\|_2.
\label{eq:app_hadamard_bound_short}
\end{equation}

Using the submultiplicativity of the spectral norm, Eq.~\ref{eq:app_swiglu_forward} satisfies

\begin{equation}
\begin{aligned}
\left\|
f_{l,e}(x)
\right\|_2
&\leq
\left\|
\operatorname{SiLU}(u)\odot v
\right\|_2
\left\|
W_{l,e}^{\mathrm{down}}
\right\|_2
\\
&\leq
\left\|
\operatorname{SiLU}(u)
\right\|_2
\|v\|_2
\left\|
W_{l,e}^{\mathrm{down}}
\right\|_2
\\
&\leq
\|u\|_2
\|v\|_2
\left\|
W_{l,e}^{\mathrm{down}}
\right\|_2.
\end{aligned}
\label{eq:app_swiglu_bound_intermediate}
\end{equation}

The two branch responses can be further bounded as

\begin{equation}
\begin{aligned}
\|u\|_2
&=
\left\|
\bar{x}\Gamma_lW_{l,e}^{\mathrm{gate}}
\right\|_2
\\
&\leq
\|\bar{x}\|_2
\left\|
\Gamma_lW_{l,e}^{\mathrm{gate}}
\right\|_2,
\\
\|v\|_2
&=
\left\|
\bar{x}\Gamma_lW_{l,e}^{\mathrm{up}}
\right\|_2
\\
&\leq
\|\bar{x}\|_2
\left\|
\Gamma_lW_{l,e}^{\mathrm{up}}
\right\|_2.
\end{aligned}
\label{eq:app_branch_response_bounds}
\end{equation}

Substituting Eq.~\ref{eq:app_branch_response_bounds} into Eq.~\ref{eq:app_swiglu_bound_intermediate} yields

\begin{equation}
\boxed{
\left\|
f_{l,e}(x)
\right\|_2
\leq
\|\bar{x}\|_2^2
\left\|
\Gamma_lW_{l,e}^{\mathrm{gate}}
\right\|_2
\left\|
\Gamma_lW_{l,e}^{\mathrm{up}}
\right\|_2
\left\|
W_{l,e}^{\mathrm{down}}
\right\|_2
}.
\label{eq:app_swiglu_final_bound}
\end{equation}

This result shows that the potential response of a SwiGLU expert depends jointly and multiplicatively on the gate branch, the up branch, and the down projection. Therefore, a proxy based on only one projection matrix cannot fully capture the structural response capacity of the expert. The branch-symmetric construction of GSP preserves all three components through two complementary factorizations of this multiplicative dependence.

\subsubsection{Shared-Translation Invariance of Router Weights}
\label{app:router_translation}

We prove that softmax routing probabilities are invariant to a shared translation of all router weight vectors. This property motivates centering the router weights before constructing expert-specific directional prototypes.

For the $l$-th MoE layer, let the routing logit of expert $e$ be

\begin{equation}
z_{l,e}(x)
=
x^{\top}w_{l,e}+b_{l,e}.
\label{eq:app_router_logit}
\end{equation}

The corresponding softmax routing probability is

\begin{equation}
\pi_{l,e}(x)
=
\frac{
\exp\left(z_{l,e}(x)\right)
}{
\sum_{j=1}^{N_l}
\exp\left(z_{l,j}(x)\right)
}.
\label{eq:app_router_probability}
\end{equation}

Consider adding a common vector $c_l$ to every router weight vector:

\begin{equation}
w_{l,e}'
=
w_{l,e}+c_l,
\qquad
e=1,\ldots,N_l.
\label{eq:app_router_translation}
\end{equation}

The translated routing logit becomes

\begin{equation}
\begin{aligned}
z_{l,e}'(x)
&=
x^{\top}w_{l,e}'+b_{l,e}
\\
&=
x^{\top}w_{l,e}
+
x^{\top}c_l
+
b_{l,e}
\\
&=
z_{l,e}(x)+x^{\top}c_l.
\end{aligned}
\label{eq:app_translated_logit}
\end{equation}

Because the additive term $x^{\top}c_l$ is identical for all experts, the resulting routing probability is

\begin{equation}
\begin{aligned}
\pi_{l,e}'(x)
&=
\frac{
\exp\left(z_{l,e}(x)+x^{\top}c_l\right)
}{
\sum_{j=1}^{N_l}
\exp\left(z_{l,j}(x)+x^{\top}c_l\right)
}
\\
&=
\frac{
\exp\left(x^{\top}c_l\right)
\exp\left(z_{l,e}(x)\right)
}{
\exp\left(x^{\top}c_l\right)
\sum_{j=1}^{N_l}
\exp\left(z_{l,j}(x)\right)
}
\\
&=
\pi_{l,e}(x).
\end{aligned}
\label{eq:app_router_probability_invariance}
\end{equation}

Therefore, simultaneously translating all router weights by the same vector does not alter the routing probabilities. The same argument also applies to adding a shared scalar to all router biases.

We next show that the centered router representation is invariant to this shared translation. Define the layer-wise mean router weight as

\begin{equation}
\overline{w}_l
=
\frac{1}{N_l}
\sum_{j=1}^{N_l}
w_{l,j},
\label{eq:app_router_mean}
\end{equation}

and the centered weight of expert $e$ as

\begin{equation}
\widetilde{w}_{l,e}
=
w_{l,e}-\overline{w}_l.
\label{eq:app_centered_router_weight}
\end{equation}

After the shared translation in Eq.~\ref{eq:app_router_translation}, the layer-wise mean becomes

\begin{equation}
\begin{aligned}
\overline{w}_l'
&=
\frac{1}{N_l}
\sum_{j=1}^{N_l}
\left(
w_{l,j}+c_l
\right)
\\
&=
\overline{w}_l+c_l.
\end{aligned}
\label{eq:app_translated_router_mean}
\end{equation}

The centered translated weight is therefore

\begin{equation}
\begin{aligned}
\widetilde{w}_{l,e}'
&=
w_{l,e}'-\overline{w}_l'
\\
&=
\left(
w_{l,e}+c_l
\right)
-
\left(
\overline{w}_l+c_l
\right)
\\
&=
w_{l,e}-\overline{w}_l
\\
&=
\widetilde{w}_{l,e}.
\end{aligned}
\label{eq:app_centering_invariance}
\end{equation}

Hence, $\widetilde{w}_{l,e}$ removes the non-identifiable shared component of the router weights and preserves only the direction that discriminates expert $e$ relative to the other experts in the same layer. This makes the centered router weight a more appropriate representation of expert-specific routing preference than the uncentered weight.

\subsubsection{Conditional Mean Shift Under Gaussian Exponential Tilting}
\label{app:gaussian_tilting}

We derive the conditional mean shift induced by the centered router direction under a Gaussian approximation. This result provides the distributional interpretation of $\widetilde{w}_{l,e}$ as an expert-conditioned input direction.

Let the hidden-state distribution at the $l$-th MoE layer be approximated by

\begin{equation}
x
\sim
\mathcal{N}
\left(
\mu_l^x,
\Sigma_l
\right).
\label{eq:app_hidden_gaussian}
\end{equation}

For compactness, let $w=\widetilde{w}_{l,e}$, $\mu=\mu_l^x$, and $\Sigma=\Sigma_l$. The Gaussian density can be written as

\begin{equation}
p_l(x)
=
\frac{
1
}{
(2\pi)^{d/2}
|\Sigma|^{1/2}
}
\exp
\left(
-\frac{1}{2}
(x-\mu)^{\top}
\Sigma^{-1}
(x-\mu)
\right).
\label{eq:app_gaussian_density}
\end{equation}

The exponentially tilted distribution induced by $w$ is

\begin{equation}
p_{l,e}(x)
=
\frac{
\exp(w^{\top}x)p_l(x)
}{
\mathbb{E}_{x\sim p_l}
\left[
\exp(w^{\top}x)
\right]
}.
\label{eq:app_tilted_distribution}
\end{equation}

The exponent in the unnormalized density is

\begin{equation}
-\frac{1}{2}
(x-\mu)^{\top}
\Sigma^{-1}
(x-\mu)
+
w^{\top}x.
\label{eq:app_tilted_exponent}
\end{equation}

Expanding the quadratic term gives

\begin{equation}
\begin{aligned}
&
-\frac{1}{2}
x^{\top}\Sigma^{-1}x
+
x^{\top}\Sigma^{-1}\mu
-
\frac{1}{2}
\mu^{\top}\Sigma^{-1}\mu
+
w^{\top}x
\\
&=
-\frac{1}{2}
x^{\top}\Sigma^{-1}x
+
x^{\top}
\Sigma^{-1}
\left(
\mu+\Sigma w
\right)
-
\frac{1}{2}
\mu^{\top}\Sigma^{-1}\mu.
\end{aligned}
\label{eq:app_tilted_expansion}
\end{equation}

Completing the square with respect to $x$ yields

\begin{equation}
\begin{aligned}
&
-\frac{1}{2}
(x-\mu)^{\top}
\Sigma^{-1}
(x-\mu)
+
w^{\top}x
\\
&=
-\frac{1}{2}
\left(
x-\mu-\Sigma w
\right)^{\top}
\Sigma^{-1}
\left(
x-\mu-\Sigma w
\right)
\\
&\quad
+
w^{\top}\mu
+
\frac{1}{2}
w^{\top}\Sigma w.
\end{aligned}
\label{eq:app_complete_square}
\end{equation}

The last two terms do not depend on $x$ and are absorbed into the normalization constant. In particular, the moment-generating function of the Gaussian distribution gives

\begin{equation}
\mathbb{E}_{x\sim p_l}
\left[
\exp(w^{\top}x)
\right]
=
\exp
\left(
w^{\top}\mu
+
\frac{1}{2}
w^{\top}\Sigma w
\right).
\label{eq:app_gaussian_mgf}
\end{equation}

It follows that the tilted distribution is also Gaussian:

\begin{equation}
p_{l,e}(x)
=
\mathcal{N}
\left(
\mu+\Sigma w,
\Sigma
\right).
\label{eq:app_tilted_gaussian}
\end{equation}

Therefore, its conditional mean satisfies

\begin{equation}
\mathbb{E}_{x\sim p_{l,e}}[x]
=
\mu_l^x
+
\Sigma_l\widetilde{w}_{l,e},
\label{eq:app_tilted_mean}
\end{equation}

and the mean shift relative to the original hidden-state distribution is

\begin{equation}
\boxed{
\mathbb{E}_{x\sim p_{l,e}}[x]
-
\mathbb{E}_{x\sim p_l}[x]
=
\Sigma_l\widetilde{w}_{l,e}
}.
\label{eq:app_conditional_mean_shift}
\end{equation}

When the hidden-state covariance is approximately isotropic, i.e., $\Sigma_l\approx\sigma_l^2I$, this result becomes

\begin{equation}
\mathbb{E}_{x\sim p_{l,e}}[x]
-
\mathbb{E}_{x\sim p_l}[x]
\approx
\sigma_l^2\widetilde{w}_{l,e}.
\label{eq:app_isotropic_mean_shift}
\end{equation}

Thus, the conditional mean-shift direction is approximately aligned with the centered router weight. RCR consequently uses the RMS-normalized version of $\widetilde{w}_{l,e}$ as a data-free approximation to the characteristic input direction associated with expert $e$.

\subsubsection{Conservativeness and Output Perturbation of Dual-View Skipping}
\label{app:ace_conservative}

We prove that the maximum-based ACE criterion is equivalent to intersecting the low-contribution candidate sets identified by GSP and RCR. We further derive an output-perturbation bound for the resulting expert skipping and gate renormalization.

\paragraph{Equivalence to candidate-set intersection.}
For a routed expert slot $i$, ACE defines

\begin{equation}
c_{l,t,i}^{\mathrm{ACE}}
=
\max
\left(
p_{l,t,i}^{\mathrm{GSP}},
p_{l,t,i}^{\mathrm{RCR}}
\right).
\label{eq:app_ace_max_score}
\end{equation}

Given a threshold $\tau_q=\tau(q)$, the expert slot is initially considered skippable if

\begin{equation}
c_{l,t,i}^{\mathrm{ACE}}
<
\tau_q.
\label{eq:app_ace_skip_condition}
\end{equation}

For any scalars $a$, $b$, and $\tau$, the following equivalence holds:

\begin{equation}
\max(a,b)<\tau
\quad\Longleftrightarrow\quad
a<\tau
\ \text{and}\
b<\tau.
\label{eq:app_max_equivalence}
\end{equation}

Applying Eq.~\ref{eq:app_max_equivalence} gives

\begin{equation}
\begin{aligned}
c_{l,t,i}^{\mathrm{ACE}}<\tau_q
\quad\Longleftrightarrow\quad
&
p_{l,t,i}^{\mathrm{GSP}}<\tau_q
\\
&\text{and}\quad
p_{l,t,i}^{\mathrm{RCR}}<\tau_q.
\end{aligned}
\label{eq:app_ace_slot_equivalence}
\end{equation}

Define the two low-contribution candidate sets as

\begin{equation}
\begin{aligned}
\mathcal{S}_{l,t}^{\mathrm{GSP}}
&=
\left\{
i
\;\middle|\;
p_{l,t,i}^{\mathrm{GSP}}<\tau_q
\right\},
\\
\mathcal{S}_{l,t}^{\mathrm{RCR}}
&=
\left\{
i
\;\middle|\;
p_{l,t,i}^{\mathrm{RCR}}<\tau_q
\right\}.
\end{aligned}
\label{eq:app_dual_candidate_sets}
\end{equation}

Equation~\ref{eq:app_ace_slot_equivalence} directly implies

\begin{equation}
\boxed{
\mathcal{S}_{l,t}^{\mathrm{ACE}}
=
\mathcal{S}_{l,t}^{\mathrm{GSP}}
\cap
\mathcal{S}_{l,t}^{\mathrm{RCR}}
}.
\label{eq:app_ace_intersection}
\end{equation}

Consequently,

\begin{equation}
\mathcal{S}_{l,t}^{\mathrm{ACE}}
\subseteq
\mathcal{S}_{l,t}^{\mathrm{GSP}},
\qquad
\mathcal{S}_{l,t}^{\mathrm{ACE}}
\subseteq
\mathcal{S}_{l,t}^{\mathrm{RCR}}.
\label{eq:app_ace_subset}
\end{equation}

The subsequent top-1 and minimum-active-expert safeguards can only remove indices from the initial ACE skipping set. Hence, if $\widehat{\mathcal{S}}_{l,t}^{\mathrm{ACE}}$ denotes the actual skipping set after these safeguards, then

\begin{equation}
\widehat{\mathcal{S}}_{l,t}^{\mathrm{ACE}}
\subseteq
\mathcal{S}_{l,t}^{\mathrm{ACE}}
\subseteq
\mathcal{S}_{l,t}^{\mathrm{GSP}}
\cap
\mathcal{S}_{l,t}^{\mathrm{RCR}}.
\label{eq:app_safeguarded_subset}
\end{equation}

This proves that ACE is conservative relative to either individual view: an expert regarded as important by GSP or RCR cannot be skipped by the dual-view criterion.

\paragraph{Skipped router-mass comparison.}
Assume that the original top-$k$ gates are nonnegative and normalized such that

\begin{equation}
\sum_{i=1}^{k}g_{l,t,i}=1.
\label{eq:app_gate_normalization}
\end{equation}

For any skipping set $\mathcal{S}$, define its skipped router mass as

\begin{equation}
\alpha(\mathcal{S})
=
\sum_{i\in\mathcal{S}}
g_{l,t,i}.
\label{eq:app_skipped_mass}
\end{equation}

Since the gates are nonnegative and the ACE skipping set is a subset of either single-view set,

\begin{equation}
\begin{aligned}
\alpha
\left(
\widehat{\mathcal{S}}_{l,t}^{\mathrm{ACE}}
\right)
&\leq
\alpha
\left(
\mathcal{S}_{l,t}^{\mathrm{ACE}}
\right)
\\
&\leq
\min
\left\{
\alpha
\left(
\mathcal{S}_{l,t}^{\mathrm{GSP}}
\right),
\alpha
\left(
\mathcal{S}_{l,t}^{\mathrm{RCR}}
\right)
\right\}.
\end{aligned}
\label{eq:app_skipped_mass_comparison}
\end{equation}

Thus, under the same threshold, the dual-view rule removes no more router probability mass than either individual-view rule.

\paragraph{Output perturbation after gate renormalization.}
For readability, omit the layer and token indices and let $\mathcal{S}$ and $\mathcal{I}$ denote the skipped and retained expert-slot sets, respectively. Define

\begin{equation}
\alpha
=
\sum_{i\in\mathcal{S}}g_i,
\qquad
1-\alpha
=
\sum_{i\in\mathcal{I}}g_i.
\label{eq:app_alpha_definition}
\end{equation}

The original MoE output is

\begin{equation}
y
=
\sum_{i\in\mathcal{I}}g_if_i(x)
+
\sum_{i\in\mathcal{S}}g_if_i(x).
\label{eq:app_original_moe_output}
\end{equation}

After skipping and renormalization, the retained gate is $\widehat{g}_i=g_i/(1-\alpha)$, and the ACE output becomes

\begin{equation}
\widehat{y}
=
\sum_{i\in\mathcal{I}}
\frac{g_i}{1-\alpha}
f_i(x).
\label{eq:app_skipped_moe_output}
\end{equation}

For $0<\alpha<1$, define the gate-weighted mean responses of the retained and skipped experts as

\begin{equation}
\begin{aligned}
\mu_{\mathcal{I}}
&=
\frac{1}{1-\alpha}
\sum_{i\in\mathcal{I}}
g_if_i(x),
\\
\mu_{\mathcal{S}}
&=
\frac{1}{\alpha}
\sum_{i\in\mathcal{S}}
g_if_i(x).
\end{aligned}
\label{eq:app_group_mean_outputs}
\end{equation}

The original and sparsified outputs can then be written as

\begin{equation}
y
=
(1-\alpha)\mu_{\mathcal{I}}
+
\alpha\mu_{\mathcal{S}},
\qquad
\widehat{y}
=
\mu_{\mathcal{I}}.
\label{eq:app_output_mean_form}
\end{equation}

Their difference is therefore

\begin{equation}
y-\widehat{y}
=
\alpha
\left(
\mu_{\mathcal{S}}
-
\mu_{\mathcal{I}}
\right).
\label{eq:app_output_difference}
\end{equation}

Taking the Euclidean norm gives

\begin{equation}
\boxed{
\left\|
y-\widehat{y}
\right\|_2
=
\alpha
\left\|
\mu_{\mathcal{S}}
-
\mu_{\mathcal{I}}
\right\|_2
}.
\label{eq:app_exact_perturbation}
\end{equation}

This identity shows that the output perturbation is jointly determined by the skipped router mass $\alpha$ and the discrepancy between the average responses of the skipped and retained experts.

Suppose that the expert outputs satisfy

\begin{equation}
\|f_i(x)\|_2
\leq
F_{\max}
\qquad
\text{for all }i\in\{1,\ldots,k\}.
\label{eq:app_expert_output_bound}
\end{equation}

Because $\mu_{\mathcal{I}}$ and $\mu_{\mathcal{S}}$ are convex combinations of expert outputs,

\begin{equation}
\left\|
\mu_{\mathcal{I}}
\right\|_2
\leq
F_{\max},
\qquad
\left\|
\mu_{\mathcal{S}}
\right\|_2
\leq
F_{\max}.
\label{eq:app_group_output_bounds}
\end{equation}

Applying the triangle inequality to Eq.~\ref{eq:app_exact_perturbation} yields

\begin{equation}
\boxed{
\left\|
y-\widehat{y}
\right\|_2
\leq
2\alpha F_{\max}
}.
\label{eq:app_output_perturbation_bound}
\end{equation}

For the degenerate case $\alpha=0$, no expert is skipped and the perturbation is exactly zero. The top-1 retention rule guarantees that $\alpha<1$, so the gate-renormalized output remains well-defined.

Combining Eq.~\ref{eq:app_skipped_mass_comparison} with Eq.~\ref{eq:app_output_perturbation_bound} shows that the worst-case perturbation bound of ACE is no larger than that induced by either single-view candidate set under the same threshold:

\begin{equation}
\begin{aligned}
\left\|
y-\widehat{y}_{\mathrm{ACE}}
\right\|_2
&\leq
2F_{\max}
\alpha
\left(
\widehat{\mathcal{S}}_{l,t}^{\mathrm{ACE}}
\right)
\\
&\leq
2F_{\max}
\min
\left\{
\alpha
\left(
\mathcal{S}_{l,t}^{\mathrm{GSP}}
\right),
\alpha
\left(
\mathcal{S}_{l,t}^{\mathrm{RCR}}
\right)
\right\}.
\end{aligned}
\label{eq:app_ace_perturbation_comparison}
\end{equation}

Therefore, the dual-view intersection, top-1 retention, and minimum-active-expert constraint jointly provide a conservative safeguard against excessive output perturbation, while still allowing token-adaptive expert sparsification.

\section{Additional Experimental Details}
\label{app:experiments}

\subsection{Budget-to-Threshold Mapping}
\label{app:threshold_mapping}

ACE derives both expert-importance estimators exclusively from frozen model parameters. Neither GSP nor RCR uses calibration samples, evaluation inputs, labels, losses, or parameter updates. Deploying a requested execution budget nevertheless requires expressing that budget in the numerical scale of the data-free scores. We perform this separate budget-control step with one unlabeled score-collection pass. Let $N_{\mathrm{all}}$ be the number of routed expert slots observed in the pass and $N_{\mathrm{cand}}$ the number remaining after excluding the slots protected by the top-1 and minimum-keep constraints. For a requested global skipping ratio $q$, the number of skipped slots is
\begin{equation}
m_q=\operatorname{clip}\!\left(\operatorname{round}(qN_{\mathrm{all}}),0,N_{\mathrm{cand}}\right).
\end{equation}
If the sorted candidate scores are $c_{(1)}\leq\cdots\leq c_{(N_{\mathrm{cand}})}$, we place $\tau_q$ between $c_{(m_q)}$ and $c_{(m_q+1)}$ whenever the two values differ. Tied boundary values are handled by choosing the adjacent representable threshold whose realized count is closest to $m_q$. One collected score distribution therefore yields the complete threshold table for all budgets.

This procedure is deliberately separated from expert scoring and does not alter the GSP or RCR lookup tables. Its only role is budget control. After accounting for safeguarded slots and boundary ties, we report the measured fraction of skipped routed slots; hence, the 10--60\% operating points in the paper are realized rather than merely requested ratios.

Table~\ref{tab:threshold_stability} reports the ACE thresholds obtained independently from every evaluation workload. For a fixed model and realized skipping ratio, the thresholds remain tightly clustered across datasets despite their different domains and output formats. This consistency shows that the score scale is primarily determined by the pretrained model and the requested execution budget rather than by a specific benchmark. Consequently, a threshold obtained from one unlabeled workload can be reused on another without recomputing GSP or RCR, supporting cross-dataset deployment of the data-free contribution estimators.

\input{Tab/threshold_stability}

\subsection{Complete Task-Level Results}
\label{app:complete_task_results}

Tables~\ref{tab:full_qwen3_10}--\ref{tab:full_gemma_60} report every task-level result underlying the budget curves in the main paper. Task metrics and Avg. are accuracies (\%); PPL is evaluated on WikiText-2. Within each model and skipping budget, BF16 is an unpruned reference and is excluded from ranking. Boldface marks the best result among all pruning methods for each metric, with ties retained. Tables are ordered by model and then by target expert-skipping ratio from 10\% to 60\%.

\input{Tab/full_qwen3_10}
\input{Tab/full_qwen3_20}
\input{Tab/full_qwen3_30}
\input{Tab/full_qwen3_40}
\input{Tab/full_qwen3_50}
\input{Tab/full_qwen3_60}
\input{Tab/full_qwen36_10}
\input{Tab/full_qwen36_20}
\input{Tab/full_qwen36_30}
\input{Tab/full_qwen36_40}
\input{Tab/full_qwen36_50}
\input{Tab/full_qwen36_60}
\input{Tab/full_gemma_10}
\input{Tab/full_gemma_20}
\input{Tab/full_gemma_30}
\input{Tab/full_gemma_40}
\input{Tab/full_gemma_50}
\input{Tab/full_gemma_60}

%% file: Tab/threshold_stability.tex
\begin{table*}[t]
\caption{Cross-workload stability of ACE thresholds. Each cell reports the threshold independently obtained from the corresponding unlabeled evaluation workload at the specified realized skipping ratio.}
\label{tab:threshold_stability}
\centering
\scriptsize
\setlength{\tabcolsep}{3.0pt}
\renewcommand{\arraystretch}{0.96}
\begin{tabular}{llrrrrrr}
\toprule
Model & Workload & 10\% & 20\% & 30\% & 40\% & 50\% & 60\% \\
\midrule
\multirow{8}{*}{Qwen3-30B-A3B}
& WikiText-2 & 0.066076 & 0.078141 & 0.088436 & 0.098972 & 0.110965 & 0.125460 \\
& ARC-C & 0.058638 & 0.071791 & 0.082505 & 0.093486 & 0.105523 & 0.121092 \\
& ARC-E & 0.058372 & 0.071509 & 0.082139 & 0.093135 & 0.105159 & 0.120699 \\
& PIQA & 0.059348 & 0.073267 & 0.084159 & 0.095478 & 0.106510 & 0.121876 \\
& MATH-500 & 0.064817 & 0.077139 & 0.087528 & 0.097363 & 0.108013 & 0.120972 \\
& GPQA-Diamond & 0.062387 & 0.074762 & 0.085411 & 0.095339 & 0.106186 & 0.120229 \\
& HumanEval & 0.059915 & 0.073406 & 0.084152 & 0.094608 & 0.105755 & 0.119866 \\
& LiveCodeBench & 0.063674 & 0.076165 & 0.086691 & 0.096949 & 0.107934 & 0.121980 \\
\midrule
\multirow{8}{*}{Qwen3.6-35B-A3B}
& WikiText-2 & 0.068089 & 0.080015 & 0.089632 & 0.098947 & 0.109093 & 0.121344 \\
& ARC-C & 0.058495 & 0.070524 & 0.080678 & 0.091051 & 0.102505 & 0.116258 \\
& ARC-E & 0.058061 & 0.070070 & 0.080227 & 0.090632 & 0.102167 & 0.116018 \\
& PIQA & 0.058142 & 0.071141 & 0.081325 & 0.091348 & 0.102849 & 0.116473 \\
& MATH-500 & 0.060603 & 0.073996 & 0.084784 & 0.095119 & 0.105744 & 0.117395 \\
& GPQA-Diamond & 0.066227 & 0.078559 & 0.088131 & 0.097670 & 0.107836 & 0.119644 \\
& HumanEval & 0.060916 & 0.075139 & 0.086014 & 0.096280 & 0.106390 & 0.117196 \\
& LiveCodeBench & 0.062504 & 0.076356 & 0.086684 & 0.096633 & 0.106850 & 0.118335 \\
\midrule
\multirow{9}{*}{Gemma-4-26B-A4B-it}
& WikiText-2 & 0.051152 & 0.068149 & 0.081192 & 0.093195 & 0.105726 & 0.120364 \\
& AIME 2025 & 0.054991 & 0.070915 & 0.083559 & 0.095452 & 0.108153 & 0.122674 \\
& ARC-C & 0.052549 & 0.069626 & 0.082656 & 0.094467 & 0.107235 & 0.121681 \\
& ARC-E & 0.051525 & 0.068566 & 0.081766 & 0.093692 & 0.106557 & 0.121260 \\
& PIQA & 0.054232 & 0.071812 & 0.084943 & 0.096504 & 0.108424 & 0.122556 \\
& MATH-500 & 0.056684 & 0.073440 & 0.085783 & 0.097104 & 0.109174 & 0.122829 \\
& GPQA-Diamond & 0.055609 & 0.072302 & 0.084753 & 0.096559 & 0.108750 & 0.122434 \\
& HumanEval & 0.057597 & 0.073689 & 0.085818 & 0.097168 & 0.109297 & 0.122688 \\
& LiveCodeBench & 0.056699 & 0.073054 & 0.085367 & 0.096766 & 0.108862 & 0.122271 \\
\bottomrule
\end{tabular}
\end{table*}

%% file: Tab/full_qwen3_10.tex
\begin{table}[t]
\caption{Qwen3-30B-A3B-Instruct-2507 at 10\% expert skipping.}
\label{tab:full_qwen3_10}
\centering
\tablestyle{2.0pt}{1.02}
\resizebox{\columnwidth}{!}{%
\begin{tabular}{lrrrrrrrrr}
\toprule
Method & PPL$\downarrow$ & ARC-C & ARC-E & GPQA & HumanEval & LiveCodeBench & MATH-500 & PIQA & Avg. \\
\midrule
BF16 & 7.52 & 93.17 & 94.23 & 56.57 & 93.90 & 55.26 & 80.00 & 91.57 & 80.67 \\
\midrule
Score & 7.70 & 93.00 & 94.36 & 53.03 & 93.29 & 52.17 & \textbf{80.00} & \textbf{91.95} & 79.69 \\
NAEE & 7.70 & \textbf{93.33} & 94.27 & \textbf{57.57} & 92.94 & \textbf{53.37} & \textbf{80.00} & 91.22 & \textbf{80.39} \\
MoDES & 7.57 & 93.10 & 94.29 & 55.78 & 94.10 & 50.23 & 79.80 & 91.20 & 79.79 \\
DiEP & 10.09 & 90.78 & 93.98 & 48.48 & 86.59 & 48.72 & 78.60 & 88.90 & 76.58 \\
AIMER & 7.55 & 93.09 & 94.32 & 55.97 & 94.10 & 53.11 & 78.60 & 91.35 & 80.08 \\
Top-$P$ & \textbf{7.54} & 93.23 & 94.23 & 55.05 & 94.11 & 53.18 & 79.00 & 90.86 & 79.95 \\
SERE & 8.77 & 91.55 & 93.77 & 52.22 & 78.48 & 50.86 & 75.40 & 86.83 & 75.59 \\
XShare & 8.10 & 91.72 & 91.11 & 53.31 & 82.01 & 50.40 & 75.80 & 89.33 & 76.24 \\
ExpertSparsity & 7.68 & 92.83 & 94.32 & 55.01 & 92.48 & 52.37 & 78.40 & 91.40 & 79.54 \\
GSP & 7.56 & 93.19 & 94.31 & 55.77 & 94.33 & 53.27 & 78.80 & 91.29 & 80.14 \\
RCR & 7.59 & 93.17 & 94.23 & 55.05 & 93.90 & 53.18 & 79.60 & 91.78 & 80.13 \\
ACE & 7.55 & 93.26 & \textbf{94.40} & 56.06 & \textbf{94.51} & 53.27 & 79.80 & 91.40 & \textbf{80.39} \\
\bottomrule
\end{tabular}
}
\end{table}

%% file: Tab/full_qwen3_20.tex
\begin{table}[t]
\caption{Qwen3-30B-A3B-Instruct-2507 at 20\% expert skipping.}
\label{tab:full_qwen3_20}
\centering
\tablestyle{2.0pt}{1.02}
\resizebox{\columnwidth}{!}{%
\begin{tabular}{lrrrrrrrrr}
\toprule
Method & PPL$\downarrow$ & ARC-C & ARC-E & GPQA & HumanEval & LiveCodeBench & MATH-500 & PIQA & Avg. \\
\midrule
BF16 & 7.52 & 93.17 & 94.23 & 56.57 & 93.90 & 55.26 & 80.00 & 91.57 & 80.67 \\
\midrule
Score & 8.09 & 93.00 & 94.23 & 48.99 & 91.46 & 48.18 & 78.20 & 90.86 & 77.85 \\
NAEE & 8.07 & 92.83 & \textbf{94.36} & 52.02 & \textbf{92.34} & 52.89 & \textbf{80.00} & 90.96 & 79.34 \\
MoDES & 7.84 & \textbf{93.19} & 94.23 & 54.11 & 91.48 & 50.10 & 78.40 & 90.10 & 78.80 \\
DiEP & 16.00 & 87.29 & 91.41 & 19.19 & 60.98 & 47.39 & 59.80 & 84.44 & 64.36 \\
AIMER & \textbf{7.65} & 93.09 & 94.23 & 57.10 & 91.46 & 54.17 & 78.80 & 90.70 & 79.94 \\
Top-$P$ & 7.66 & 93.13 & 94.23 & 56.42 & 91.64 & 52.76 & 78.20 & 90.42 & 79.54 \\
SERE & 11.00 & 89.25 & 92.47 & 25.83 & 68.31 & 47.33 & 60.80 & 82.59 & 66.65 \\
XShare & 9.14 & 73.98 & 80.12 & 40.77 & 70.36 & 38.67 & 58.80 & 80.17 & 63.27 \\
ExpertSparsity & 8.02 & 93.00 & 94.28 & 56.28 & 90.66 & 52.39 & 78.60 & 90.91 & 79.45 \\
GSP & 7.67 & 92.77 & 94.23 & 56.90 & 91.76 & 54.29 & 78.40 & \textbf{91.21} & 79.94 \\
RCR & 7.83 & 92.41 & 94.23 & 54.04 & 91.46 & 50.17 & 79.40 & 90.26 & 78.85 \\
ACE & 7.66 & 93.09 & 94.23 & \textbf{58.08} & 92.26 & \textbf{54.50} & 78.60 & 91.02 & \textbf{80.25} \\
\bottomrule
\end{tabular}
}
\end{table}

%% file: Tab/full_qwen3_30.tex
\begin{table}[t]
\caption{Qwen3-30B-A3B-Instruct-2507 at 30\% expert skipping.}
\label{tab:full_qwen3_30}
\centering
\tablestyle{2.0pt}{1.02}
\resizebox{\columnwidth}{!}{%
\begin{tabular}{lrrrrrrrrr}
\toprule
Method & PPL$\downarrow$ & ARC-C & ARC-E & GPQA & HumanEval & LiveCodeBench & MATH-500 & PIQA & Avg. \\
\midrule
BF16 & 7.52 & 93.17 & 94.23 & 56.57 & 93.90 & 55.26 & 80.00 & 91.57 & 80.67 \\
\midrule
Score & 8.65 & 92.58 & 94.28 & 50.51 & 92.68 & 48.01 & 78.20 & 89.55 & 77.97 \\
NAEE & 8.68 & 91.89 & 94.11 & 50.00 & 91.29 & 47.20 & 79.00 & \textbf{90.91} & 77.77 \\
MoDES & 8.33 & 91.19 & \textbf{94.89} & 51.98 & 93.01 & 49.88 & 75.40 & 89.60 & 77.99 \\
DiEP & 28.42 & 78.41 & 86.78 & 4.55 & 47.56 & 37.86 & 3.80 & 70.40 & 47.05 \\
AIMER & \textbf{7.82} & \textbf{92.66} & 94.23 & 56.09 & 93.14 & 51.77 & 79.00 & 88.66 & 79.36 \\
Top-$P$ & \textbf{7.82} & 92.43 & 93.40 & 54.39 & 93.11 & 52.48 & 79.00 & 90.10 & 79.27 \\
SERE & 14.27 & 83.87 & 91.33 & 26.44 & 62.09 & 40.78 & 56.20 & 77.15 & 62.55 \\
XShare & 10.82 & 26.62 & 36.83 & 33.28 & 60.82 & 30.78 & 20.20 & 22.21 & 32.96 \\
ExpertSparsity & 8.49 & 92.49 & 91.71 & 54.29 & 92.86 & 52.77 & 77.40 & 88.90 & 78.63 \\
GSP & 7.87 & 92.01 & 94.22 & 56.42 & \textbf{93.39} & \textbf{53.20} & 78.60 & 89.28 & 79.59 \\
RCR & 8.31 & 91.64 & 94.19 & 52.02 & 92.07 & 49.67 & 75.40 & 89.66 & 77.81 \\
ACE & 7.86 & \textbf{92.66} & 94.36 & \textbf{56.56} & 93.29 & 52.66 & \textbf{79.40} & 89.45 & \textbf{79.77} \\
\bottomrule
\end{tabular}
}
\end{table}

%% file: Tab/full_qwen3_40.tex
\begin{table}[t]
\caption{Qwen3-30B-A3B-Instruct-2507 at 40\% expert skipping.}
\label{tab:full_qwen3_40}
\centering
\tablestyle{2.0pt}{1.02}
\resizebox{\columnwidth}{!}{%
\begin{tabular}{lrrrrrrrrr}
\toprule
Method & PPL$\downarrow$ & ARC-C & ARC-E & GPQA & HumanEval & LiveCodeBench & MATH-500 & PIQA & Avg. \\
\midrule
BF16 & 7.52 & 93.17 & 94.23 & 56.57 & 93.90 & 55.26 & 80.00 & 91.57 & 80.67 \\
\midrule
Score & 9.63 & 89.27 & 93.64 & 49.49 & 83.93 & 46.92 & 76.60 & 87.87 & 75.39 \\
NAEE & 9.82 & 89.51 & 93.35 & 38.89 & 83.15 & 37.82 & \textbf{79.00} & 88.82 & 72.93 \\
MoDES & 9.28 & 89.17 & 93.30 & 48.99 & 85.95 & 49.91 & 76.40 & 88.48 & 76.03 \\
DiEP & 65.87 & 54.52 & 61.11 & 5.56 & 0.00 & 22.17 & 3.80 & 44.89 & 27.44 \\
AIMER & 8.24 & 91.66 & 93.56 & 52.17 & 87.47 & 48.98 & 78.20 & 88.12 & 77.17 \\
Top-$P$ & \textbf{8.19} & 91.47 & 93.01 & 52.67 & 88.49 & 50.49 & 78.40 & \textbf{89.61} & 77.73 \\
SERE & 20.81 & 75.26 & 84.86 & 20.17 & 57.49 & 38.91 & 56.40 & 73.23 & 58.05 \\
XShare & 13.67 & 20.22 & 24.41 & 30.30 & 58.83 & 4.81 & 5.80 & 13.02 & 22.48 \\
ExpertSparsity & 9.31 & 91.47 & 93.52 & 52.00 & 86.29 & 50.10 & 76.40 & 87.54 & 76.76 \\
GSP & 8.22 & 91.88 & \textbf{94.11} & \textbf{53.10} & 88.93 & 50.29 & 78.60 & 88.14 & 77.86 \\
RCR & 9.19 & 90.70 & 93.98 & 41.92 & 86.59 & 49.10 & 76.80 & 88.41 & 75.36 \\
ACE & \textbf{8.19} & \textbf{92.15} & 94.07 & 52.02 & \textbf{90.85} & \textbf{51.37} & 78.60 & 88.41 & \textbf{78.21} \\
\bottomrule
\end{tabular}
}
\end{table}

%% file: Tab/full_qwen3_50.tex
\begin{table}[t]
\caption{Qwen3-30B-A3B-Instruct-2507 at 50\% expert skipping.}
\label{tab:full_qwen3_50}
\centering
\tablestyle{2.0pt}{1.02}
\resizebox{\columnwidth}{!}{%
\begin{tabular}{lrrrrrrrrr}
\toprule
Method & PPL$\downarrow$ & ARC-C & ARC-E & GPQA & HumanEval & LiveCodeBench & MATH-500 & PIQA & Avg. \\
\midrule
BF16 & 7.52 & 93.17 & 94.23 & 56.57 & 93.90 & 55.26 & 80.00 & 91.57 & 80.67 \\
\midrule
Score & 12.03 & 88.48 & 91.92 & 30.30 & 70.18 & 37.02 & 65.00 & 83.90 & 66.69 \\
NAEE & 12.48 & 86.43 & \textbf{92.05} & 26.77 & 67.68 & 23.89 & 73.20 & 80.75 & 64.40 \\
MoDES & 12.25 & 86.19 & 90.47 & 27.19 & 60.98 & 25.79 & 26.40 & 79.28 & 56.61 \\
DiEP & 136.92 & 5.03 & 5.39 & 3.54 & 0.00 & 5.02 & 1.80 & 6.96 & 3.96 \\
AIMER & 8.86 & 88.99 & 88.89 & 52.18 & 85.23 & 44.89 & 74.20 & 79.43 & 73.40 \\
Top-$P$ & \textbf{8.85} & 90.06 & 90.56 & 52.38 & 85.49 & 45.74 & 73.40 & 81.56 & 74.17 \\
SERE & 46.36 & 45.39 & 59.47 & 20.16 & 40.81 & 17.88 & 32.20 & 54.41 & 38.62 \\
XShare & 19.63 & 16.64 & 20.73 & 24.88 & 0.00 & 0.00 & 4.20 & 10.44 & 10.98 \\
ExpertSparsity & 10.98 & 85.32 & 91.71 & 52.18 & 86.28 & 44.28 & 72.20 & 83.73 & 73.67 \\
GSP & 8.99 & 89.79 & 90.92 & 52.49 & \textbf{87.23} & 45.89 & 73.20 & 79.16 & 74.10 \\
RCR & 11.62 & 86.26 & 91.96 & 18.69 & 77.44 & 44.81 & 65.60 & \textbf{84.33} & 67.01 \\
ACE & \textbf{8.85} & \textbf{90.78} & 89.15 & \textbf{53.54} & 86.59 & \textbf{46.26} & \textbf{74.60} & 79.16 & \textbf{74.30} \\
\bottomrule
\end{tabular}
}
\end{table}

%% file: Tab/full_qwen3_60.tex
\begin{table}[t]
\caption{Qwen3-30B-A3B-Instruct-2507 at 60\% expert skipping.}
\label{tab:full_qwen3_60}
\centering
\tablestyle{2.0pt}{1.02}
\resizebox{\columnwidth}{!}{%
\begin{tabular}{lrrrrrrrrr}
\toprule
Method & PPL$\downarrow$ & ARC-C & ARC-E & GPQA & HumanEval & LiveCodeBench & MATH-500 & PIQA & Avg. \\
\midrule
BF16 & 7.52 & 93.17 & 94.23 & 56.57 & 93.90 & 55.26 & 80.00 & 91.57 & 80.67 \\
\midrule
Score & 18.76 & 75.34 & 81.86 & 10.10 & 10.50 & 5.95 & 13.20 & 66.54 & 37.64 \\
NAEE & 19.56 & 56.25 & 56.25 & 17.68 & 27.44 & 5.88 & \textbf{60.40} & 80.21 & 43.44 \\
MoDES & 17.80 & 68.81 & 79.96 & 10.39 & 10.29 & 13.82 & 19.80 & 70.98 & 39.15 \\
DiEP & 459.88 & 4.44 & 6.40 & 4.04 & 0.00 & 0.00 & 2.60 & 7.89 & 3.62 \\
AIMER & 10.89 & 79.82 & 82.87 & 25.29 & 70.29 & 31.17 & 57.10 & 79.64 & 60.88 \\
Top-$P$ & 10.87 & 79.97 & 83.21 & 25.56 & 74.11 & 31.27 & 58.70 & 80.53 & 61.91 \\
SERE & 178.96 & 10.48 & 19.76 & 4.86 & 0.00 & 0.00 & 3.80 & 29.83 & 9.82 \\
XShare & 39.22 & 4.86 & 10.84 & 16.32 & 0.00 & 0.00 & 2.20 & 8.81 & 6.15 \\
ExpertSparsity & 13.39 & 80.42 & 83.22 & 25.18 & 74.80 & 32.92 & 59.40 & 80.02 & 62.28 \\
GSP & 10.99 & 79.61 & 83.98 & 25.93 & 74.59 & \textbf{33.59} & 59.40 & 80.58 & 62.53 \\
RCR & 17.95 & 66.81 & 80.56 & 1.52 & 17.68 & 18.80 & 23.00 & 68.82 & 39.60 \\
ACE & \textbf{10.86} & \textbf{80.97} & \textbf{85.61} & \textbf{26.77} & \textbf{75.00} & 32.61 & 60.20 & \textbf{82.32} & \textbf{63.35} \\
\bottomrule
\end{tabular}
}
\end{table}

%% file: Tab/full_qwen36_10.tex
\begin{table}[t]
\caption{Qwen3.6-35B-A3B at 10\% expert skipping.}
\label{tab:full_qwen36_10}
\centering
\tablestyle{2.0pt}{1.02}
\resizebox{\columnwidth}{!}{%
\begin{tabular}{lrrrrrrrrr}
\toprule
Method & PPL$\downarrow$ & ARC-C & ARC-E & GPQA & HumanEval & LiveCodeBench & MATH-500 & PIQA & Avg. \\
\midrule
BF16 & 7.01 & 95.14 & 94.82 & 47.47 & 95.12 & 57.35 & 84.60 & 93.69 & 81.17 \\
\midrule
Score & 7.38 & 95.31 & 94.78 & 45.45 & \textbf{97.56} & \textbf{59.72} & \textbf{84.80} & 94.02 & \textbf{81.66} \\
NAEE & 7.37 & 94.97 & \textbf{94.87} & 45.45 & 96.95 & 58.39 & 84.00 & 94.23 & 81.27 \\
MoDES & 7.38 & 95.22 & 94.39 & 47.89 & 95.31 & 56.21 & 83.20 & 93.80 & 80.86 \\
DiEP & 8.83 & 93.28 & 94.19 & 45.48 & 95.10 & 52.87 & 80.20 & 93.11 & 79.18 \\
GSP & 7.18 & 95.05 & 94.82 & 46.46 & 96.34 & 56.30 & 84.40 & \textbf{94.45} & 81.12 \\
RCR & 7.31 & 95.31 & 94.74 & \textbf{48.48} & 95.12 & 56.59 & 82.80 & 93.74 & 80.97 \\
ACE & \textbf{7.16} & \textbf{95.34} & 94.82 & \textbf{48.48} & 95.78 & 56.44 & 84.60 & 94.40 & 81.41 \\
\bottomrule
\end{tabular}
}
\end{table}

%% file: Tab/full_qwen36_20.tex
\begin{table}[t]
\caption{Qwen3.6-35B-A3B at 20\% expert skipping.}
\label{tab:full_qwen36_20}
\centering
\tablestyle{2.0pt}{1.02}
\resizebox{\columnwidth}{!}{%
\begin{tabular}{lrrrrrrrrr}
\toprule
Method & PPL$\downarrow$ & ARC-C & ARC-E & GPQA & HumanEval & LiveCodeBench & MATH-500 & PIQA & Avg. \\
\midrule
BF16 & 7.01 & 95.14 & 94.82 & 47.47 & 95.12 & 57.35 & 84.60 & 93.69 & 81.17 \\
\midrule
Score & 7.79 & 94.71 & 94.82 & 46.46 & 93.90 & 54.88 & \textbf{85.60} & 93.20 & 80.51 \\
NAEE & 7.80 & \textbf{94.88} & \textbf{94.99} & \textbf{49.49} & \textbf{96.34} & 52.89 & 84.40 & 93.96 & 80.99 \\
MoDES & 7.41 & 94.11 & 94.20 & 46.66 & 95.29 & 56.19 & 84.60 & 93.08 & 80.59 \\
DiEP & 12.09 & 91.22 & 91.17 & 42.09 & 93.67 & 49.34 & 79.80 & 91.11 & 76.91 \\
GSP & 7.38 & 93.86 & 94.82 & 47.98 & 95.73 & 51.85 & 84.80 & \textbf{94.45} & 80.50 \\
RCR & 7.65 & 94.80 & 94.78 & 45.45 & 95.73 & \textbf{58.39} & 84.60 & 93.58 & 81.05 \\
ACE & \textbf{7.34} & \textbf{94.88} & 94.82 & 47.49 & 95.73 & 58.11 & 84.80 & 94.40 & \textbf{81.46} \\
\bottomrule
\end{tabular}
}
\end{table}

%% file: Tab/full_qwen36_30.tex
\begin{table}[t]
\caption{Qwen3.6-35B-A3B at 30\% expert skipping.}
\label{tab:full_qwen36_30}
\centering
\tablestyle{2.0pt}{1.02}
\resizebox{\columnwidth}{!}{%
\begin{tabular}{lrrrrrrrrr}
\toprule
Method & PPL$\downarrow$ & ARC-C & ARC-E & GPQA & HumanEval & LiveCodeBench & MATH-500 & PIQA & Avg. \\
\midrule
BF16 & 7.01 & 95.14 & 94.82 & 47.47 & 95.12 & 57.35 & 84.60 & 93.69 & 81.17 \\
\midrule
Score & 8.36 & \textbf{95.14} & \textbf{94.78} & 46.97 & 89.02 & 47.20 & 83.40 & 92.34 & 78.41 \\
NAEE & 8.40 & \textbf{95.14} & 94.70 & 44.44 & 93.29 & 47.20 & \textbf{84.40} & \textbf{93.91} & 79.01 \\
MoDES & 7.72 & 94.77 & 94.65 & 44.79 & 89.80 & 48.19 & 84.00 & 91.11 & 78.19 \\
DiEP & 17.62 & 88.49 & 87.48 & 40.11 & 89.71 & 42.88 & 75.20 & 85.24 & 72.73 \\
GSP & 7.66 & 94.97 & 94.70 & 46.97 & 94.51 & 48.34 & 83.40 & 93.31 & 79.46 \\
RCR & 8.17 & 93.69 & 94.23 & 42.42 & 92.68 & 51.37 & 82.00 & 92.11 & 78.36 \\
ACE & \textbf{7.57} & 94.16 & 94.70 & \textbf{47.01} & \textbf{94.66} & \textbf{52.00} & 83.40 & 93.33 & \textbf{79.89} \\
\bottomrule
\end{tabular}
}
\end{table}

%% file: Tab/full_qwen36_40.tex
\begin{table}[t]
\caption{Qwen3.6-35B-A3B at 40\% expert skipping.}
\label{tab:full_qwen36_40}
\centering
\tablestyle{2.0pt}{1.02}
\resizebox{\columnwidth}{!}{%
\begin{tabular}{lrrrrrrrrr}
\toprule
Method & PPL$\downarrow$ & ARC-C & ARC-E & GPQA & HumanEval & LiveCodeBench & MATH-500 & PIQA & Avg. \\
\midrule
BF16 & 7.01 & 95.14 & 94.82 & 47.47 & 95.12 & 57.35 & 84.60 & 93.69 & 81.17 \\
\midrule
Score & 9.21 & 94.88 & 94.44 & 34.34 & 84.15 & 39.05 & 82.00 & \textbf{93.42} & 74.61 \\
NAEE & 9.30 & 94.62 & 94.40 & 38.89 & 84.15 & 37.82 & 81.20 & 92.82 & 74.84 \\
MoDES & 8.10 & 94.22 & 94.12 & 38.67 & 89.38 & 44.28 & 79.40 & 90.11 & 75.74 \\
DiEP & 28.45 & 80.84 & 82.07 & 35.75 & 80.11 & 32.98 & 71.20 & 80.16 & 66.16 \\
GSP & 8.08 & 95.05 & 94.61 & 41.41 & \textbf{95.73} & 45.21 & 82.40 & 93.36 & 78.25 \\
RCR & 8.95 & 93.94 & 94.15 & 37.88 & 88.41 & 44.74 & 78.00 & 90.32 & 75.35 \\
ACE & \textbf{7.88} & \textbf{95.10} & \textbf{94.67} & \textbf{42.56} & 95.38 & \textbf{47.88} & \textbf{82.60} & 93.33 & \textbf{78.79} \\
\bottomrule
\end{tabular}
}
\end{table}

%% file: Tab/full_qwen36_50.tex
\begin{table}[t]
\caption{Qwen3.6-35B-A3B at 50\% expert skipping.}
\label{tab:full_qwen36_50}
\centering
\tablestyle{2.0pt}{1.02}
\resizebox{\columnwidth}{!}{%
\begin{tabular}{lrrrrrrrrr}
\toprule
Method & PPL$\downarrow$ & ARC-C & ARC-E & GPQA & HumanEval & LiveCodeBench & MATH-500 & PIQA & Avg. \\
\midrule
BF16 & 7.01 & 95.14 & 94.82 & 47.47 & 95.12 & 57.35 & 84.60 & 93.69 & 81.17 \\
\midrule
Score & 10.58 & 93.77 & 94.15 & 30.81 & 60.98 & 25.21 & 75.80 & 91.46 & 67.45 \\
NAEE & 10.89 & 94.20 & 93.98 & 26.77 & 67.68 & 23.89 & 75.20 & 90.75 & 67.50 \\
MoDES & 9.42 & 93.18 & 92.65 & 33.01 & 80.49 & 34.42 & 76.00 & 90.20 & 71.42 \\
DiEP & 45.43 & 73.87 & 77.98 & 20.76 & 56.91 & 20.88 & 68.40 & 76.11 & 56.42 \\
GSP & 8.87 & 94.28 & \textbf{94.49} & 34.34 & 90.85 & 39.72 & \textbf{79.60} & \textbf{91.78} & 75.01 \\
RCR & 10.39 & 91.13 & 91.88 & 30.81 & 71.34 & 26.16 & 65.40 & 88.25 & 66.42 \\
ACE & \textbf{8.67} & \textbf{94.49} & 94.40 & \textbf{35.78} & \textbf{91.22} & \textbf{41.98} & 79.40 & 91.69 & \textbf{75.57} \\
\bottomrule
\end{tabular}
}
\end{table}

%% file: Tab/full_qwen36_60.tex
\begin{table}[t]
\caption{Qwen3.6-35B-A3B at 60\% expert skipping.}
\label{tab:full_qwen36_60}
\centering
\tablestyle{2.0pt}{1.02}
\resizebox{\columnwidth}{!}{%
\begin{tabular}{lrrrrrrrrr}
\toprule
Method & PPL$\downarrow$ & ARC-C & ARC-E & GPQA & HumanEval & LiveCodeBench & MATH-500 & PIQA & Avg. \\
\midrule
BF16 & 7.01 & 95.14 & 94.82 & 47.47 & 95.12 & 57.35 & 84.60 & 93.69 & 81.17 \\
\midrule
Score & 13.38 & 89.68 & 91.04 & 21.72 & 29.27 & 6.26 & 59.00 & 85.64 & 54.66 \\
NAEE & 14.17 & 88.31 & 91.04 & 17.68 & 27.44 & 5.88 & 61.40 & 82.21 & 53.42 \\
MoDES & 10.87 & 91.87 & 94.11 & 28.40 & 82.29 & 27.44 & 70.70 & 88.45 & 69.04 \\
DiEP & 85.12 & 70.37 & 70.11 & 11.28 & 0.00 & 0.00 & 10.20 & 63.91 & 32.27 \\
GSP & 10.48 & 93.86 & 94.28 & 30.81 & 83.54 & \textbf{27.96} & 73.20 & 90.15 & 70.54 \\
RCR & 13.48 & 78.75 & 84.18 & 18.18 & 33.54 & 4.83 & 26.80 & 71.38 & 45.38 \\
ACE & \textbf{9.98} & \textbf{93.98} & \textbf{94.30} & \textbf{31.15} & \textbf{84.20} & 27.23 & \textbf{73.80} & \textbf{90.22} & \textbf{70.70} \\
\bottomrule
\end{tabular}
}
\end{table}

%% file: Tab/full_gemma_10.tex
\begin{table}[t]
\caption{Gemma-4-26B-A4B-it at 10\% expert skipping.}
\label{tab:full_gemma_10}
\centering
\tablestyle{2.0pt}{1.02}
\resizebox{\columnwidth}{!}{%
\begin{tabular}{lrrrrrrrrr}
\toprule
Method & PPL$\downarrow$ & ARC-C & ARC-E & GPQA & HumanEval & LiveCodeBench & MATH-500 & PIQA & Avg. \\
\midrule
BF16 & -- & 95.82 & 94.91 & 51.52 & 95.73 & 59.81 & 86.60 & 87.05 & 81.63 \\
\midrule
Score & -- & 94.11 & 94.28 & 52.02 & 92.68 & 58.10 & 85.60 & 87.32 & 80.59 \\
NAEE & -- & 94.45 & \textbf{94.40} & 50.00 & 93.90 & 56.21 & \textbf{88.20} & \textbf{87.54} & 80.67 \\
MoDES & -- & 94.07 & 93.88 & 53.29 & 93.90 & 57.34 & 86.20 & 87.05 & 80.82 \\
DiEP & -- & 94.10 & 93.19 & 50.33 & 92.45 & 55.98 & 84.20 & 87.05 & 79.61 \\
GSP & -- & 94.11 & 93.90 & 53.48 & 93.90 & 57.91 & 87.40 & 87.27 & 81.14 \\
RCR & -- & 94.03 & 93.98 & 51.01 & 94.51 & 56.55 & 87.80 & 87.27 & 80.74 \\
ACE & -- & \textbf{94.87} & 93.98 & \textbf{53.65} & \textbf{95.12} & \textbf{58.29} & 88.00 & 87.21 & \textbf{81.59} \\
\bottomrule
\end{tabular}
}
\end{table}

%% file: Tab/full_gemma_20.tex
\begin{table}[t]
\caption{Gemma-4-26B-A4B-it at 20\% expert skipping.}
\label{tab:full_gemma_20}
\centering
\tablestyle{2.0pt}{1.02}
\resizebox{\columnwidth}{!}{%
\begin{tabular}{lrrrrrrrrr}
\toprule
Method & PPL$\downarrow$ & ARC-C & ARC-E & GPQA & HumanEval & LiveCodeBench & MATH-500 & PIQA & Avg. \\
\midrule
BF16 & -- & 95.82 & 94.91 & 51.52 & 95.73 & 59.81 & 86.60 & 87.05 & 81.63 \\
\midrule
Score & -- & 94.28 & 94.11 & 52.02 & 92.68 & 55.26 & 85.60 & 87.05 & 80.14 \\
NAEE & -- & 94.45 & \textbf{94.23} & \textbf{54.04} & 93.90 & 55.83 & 85.60 & 87.05 & 80.73 \\
MoDES & -- & 93.78 & 94.02 & 53.97 & 93.90 & 56.27 & 85.60 & 86.84 & 80.63 \\
DiEP & -- & 92.98 & 92.46 & 50.98 & 92.17 & 53.91 & 84.00 & 84.10 & 78.66 \\
GSP & -- & 94.11 & 94.15 & 53.54 & \textbf{95.73} & 56.94 & 86.40 & \textbf{87.27} & 81.16 \\
RCR & -- & 93.26 & 93.64 & 52.02 & 93.90 & 54.08 & 85.40 & 86.56 & 79.84 \\
ACE & -- & \textbf{94.93} & 94.17 & 53.35 & \textbf{95.73} & \textbf{57.09} & \textbf{86.80} & 86.89 & \textbf{81.28} \\
\bottomrule
\end{tabular}
}
\end{table}

%% file: Tab/full_gemma_30.tex
\begin{table}[t]
\caption{Gemma-4-26B-A4B-it at 30\% expert skipping.}
\label{tab:full_gemma_30}
\centering
\tablestyle{2.0pt}{1.02}
\resizebox{\columnwidth}{!}{%
\begin{tabular}{lrrrrrrrrr}
\toprule
Method & PPL$\downarrow$ & ARC-C & ARC-E & GPQA & HumanEval & LiveCodeBench & MATH-500 & PIQA & Avg. \\
\midrule
BF16 & -- & 95.82 & 94.91 & 51.52 & 95.73 & 59.81 & 86.60 & 87.05 & 81.63 \\
\midrule
Score & -- & 94.28 & \textbf{94.15} & 50.51 & 90.24 & 49.19 & 82.20 & 86.78 & 78.19 \\
NAEE & -- & 93.94 & 93.98 & 45.96 & 93.68 & 50.62 & 81.40 & 86.24 & 77.97 \\
MoDES & -- & 93.77 & 93.78 & 48.38 & 93.44 & 50.43 & 83.00 & 86.17 & 78.42 \\
DiEP & -- & 90.38 & 92.06 & 46.70 & 90.45 & 50.38 & 81.20 & 80.14 & 75.90 \\
GSP & -- & \textbf{94.54} & 93.98 & 52.02 & 93.34 & 54.28 & 82.80 & \textbf{87.00} & 79.71 \\
RCR & -- & 91.55 & 93.01 & 47.98 & 93.29 & 52.19 & 82.20 & 85.75 & 78.00 \\
ACE & -- & 94.44 & 94.11 & \textbf{53.03} & \textbf{93.90} & \textbf{54.77} & \textbf{83.80} & \textbf{87.00} & \textbf{80.15} \\
\bottomrule
\end{tabular}
}
\end{table}

%% file: Tab/full_gemma_40.tex
\begin{table}[t]
\caption{Gemma-4-26B-A4B-it at 40\% expert skipping.}
\label{tab:full_gemma_40}
\centering
\tablestyle{2.0pt}{1.02}
\resizebox{\columnwidth}{!}{%
\begin{tabular}{lrrrrrrrrr}
\toprule
Method & PPL$\downarrow$ & ARC-C & ARC-E & GPQA & HumanEval & LiveCodeBench & MATH-500 & PIQA & Avg. \\
\midrule
BF16 & -- & 95.82 & 94.91 & 51.52 & 95.73 & 59.81 & 86.60 & 87.05 & 81.63 \\
\midrule
Score & -- & 92.32 & 93.73 & 49.49 & 87.20 & 43.13 & 81.00 & 85.80 & 76.10 \\
NAEE & -- & 94.03 & 93.77 & 50.51 & 78.66 & 42.37 & 72.40 & 85.15 & 73.84 \\
MoDES & -- & 93.45 & 93.84 & 49.05 & 93.05 & 47.92 & 81.80 & 85.78 & 77.84 \\
DiEP & -- & 88.58 & 91.39 & 42.09 & 88.20 & 44.84 & 75.80 & 74.11 & 72.14 \\
GSP & -- & \textbf{94.45} & 94.15 & \textbf{51.01} & 93.90 & 48.98 & 81.80 & 86.29 & 78.65 \\
RCR & -- & 88.82 & 91.96 & 39.90 & 89.02 & 48.54 & 79.60 & 83.57 & 77.49 \\
ACE & -- & 94.41 & \textbf{94.44} & \textbf{51.01} & \textbf{95.12} & \textbf{52.80} & \textbf{82.20} & \textbf{86.58} & \textbf{79.51} \\
\bottomrule
\end{tabular}
}
\end{table}

%% file: Tab/full_gemma_50.tex
\begin{table}[t]
\caption{Gemma-4-26B-A4B-it at 50\% expert skipping.}
\label{tab:full_gemma_50}
\centering
\tablestyle{2.0pt}{1.02}
\resizebox{\columnwidth}{!}{%
\begin{tabular}{lrrrrrrrrr}
\toprule
Method & PPL$\downarrow$ & ARC-C & ARC-E & GPQA & HumanEval & LiveCodeBench & MATH-500 & PIQA & Avg. \\
\midrule
BF16 & -- & 95.82 & 94.91 & 51.52 & 95.73 & 59.81 & 86.60 & 87.05 & 81.63 \\
\midrule
Score & -- & 90.78 & 92.89 & 39.90 & 85.98 & 35.07 & 70.60 & 82.75 & 71.14 \\
NAEE & -- & 92.32 & 93.60 & 42.42 & 76.83 & 29.95 & 60.00 & 82.86 & 68.28 \\
MoDES & -- & 69.84 & 59.28 & 27.89 & 22.56 & 0.00 & 20.80 & 70.81 & 38.74 \\
DiEP & -- & 50.76 & 50.92 & 20.08 & 15.98 & 0.00 & 10.60 & 67.47 & 30.83 \\
GSP & -- & 93.69 & 94.19 & 46.97 & 90.85 & 41.29 & 77.40 & \textbf{85.69} & 75.73 \\
RCR & -- & 84.64 & 90.07 & 28.79 & 82.32 & 40.11 & 75.80 & 80.96 & 68.96 \\
ACE & -- & \textbf{94.11} & \textbf{94.23} & \textbf{50.00} & \textbf{92.07} & \textbf{44.89} & \textbf{78.80} & \textbf{85.69} & \textbf{77.11} \\
\bottomrule
\end{tabular}
}
\end{table}

%% file: Tab/full_gemma_60.tex
\begin{table}[t]
\caption{Gemma-4-26B-A4B-it at 60\% expert skipping.}
\label{tab:full_gemma_60}
\centering
\tablestyle{2.0pt}{1.02}
\resizebox{\columnwidth}{!}{%
\begin{tabular}{lrrrrrrrrr}
\toprule
Method & PPL$\downarrow$ & ARC-C & ARC-E & GPQA & HumanEval & LiveCodeBench & MATH-500 & PIQA & Avg. \\
\midrule
BF16 & -- & 95.82 & 94.91 & 51.52 & 95.73 & 59.81 & 86.60 & 87.05 & 81.63 \\
\midrule
Score & -- & 85.07 & 90.53 & 32.83 & 73.78 & 26.07 & 59.40 & 80.69 & 64.05 \\
NAEE & -- & 86.18 & 91.71 & 34.34 & 58.54 & 16.21 & 36.00 & 78.56 & 57.36 \\
MoDES & -- & 50.01 & 48.18 & 28.01 & 6.10 & 0.00 & 6.80 & 48.92 & 26.86 \\
DiEP & -- & 29.87 & 38.97 & 18.92 & 0.00 & 0.00 & 0.00 & 39.17 & 18.13 \\
GSP & -- & 90.53 & 93.35 & 40.40 & 87.80 & 30.98 & 61.60 & 82.86 & 69.65 \\
RCR & -- & 75.68 & 82.58 & 28.79 & 67.07 & 26.10 & 65.80 & 74.21 & 60.03 \\
ACE & -- & \textbf{91.89} & \textbf{93.73} & \textbf{41.38} & \textbf{91.46} & \textbf{33.87} & \textbf{67.60} & \textbf{84.82} & \textbf{72.11} \\
\bottomrule
\end{tabular}
}
\end{table}